\documentclass{article} 
\usepackage{iclr2026_conference,times}

\usepackage{amsmath,amsfonts,bm}

\def\eqref#1{equation~\ref{#1}}

\def\1{\bm{1}}

\DeclareMathAlphabet{\mathsfit}{\encodingdefault}{\sfdefault}{m}{sl}
\SetMathAlphabet{\mathsfit}{bold}{\encodingdefault}{\sfdefault}{bx}{n}

\usepackage{hyperref}
\usepackage{url}

\usepackage{graphicx}
\usepackage{hyperref}       
\usepackage{url}            
\usepackage{booktabs}       
\usepackage{amsfonts}       
\usepackage{nicefrac}       
\usepackage{microtype}      
\usepackage[table]{xcolor}
\definecolor{Gray}{gray}{0.9}
\usepackage{float}
\usepackage{pifont}
\usepackage{array}   
\usepackage{tabularx}   
\usepackage{adjustbox}  
\usepackage{subfig}
\usepackage[table]{xcolor}
\definecolor{Gray}{gray}{0.9}
\usepackage{caption}
\usepackage{multirow}

\title{
\begin{minipage}{0.1\textwidth}
    \includegraphics[scale=0.035]{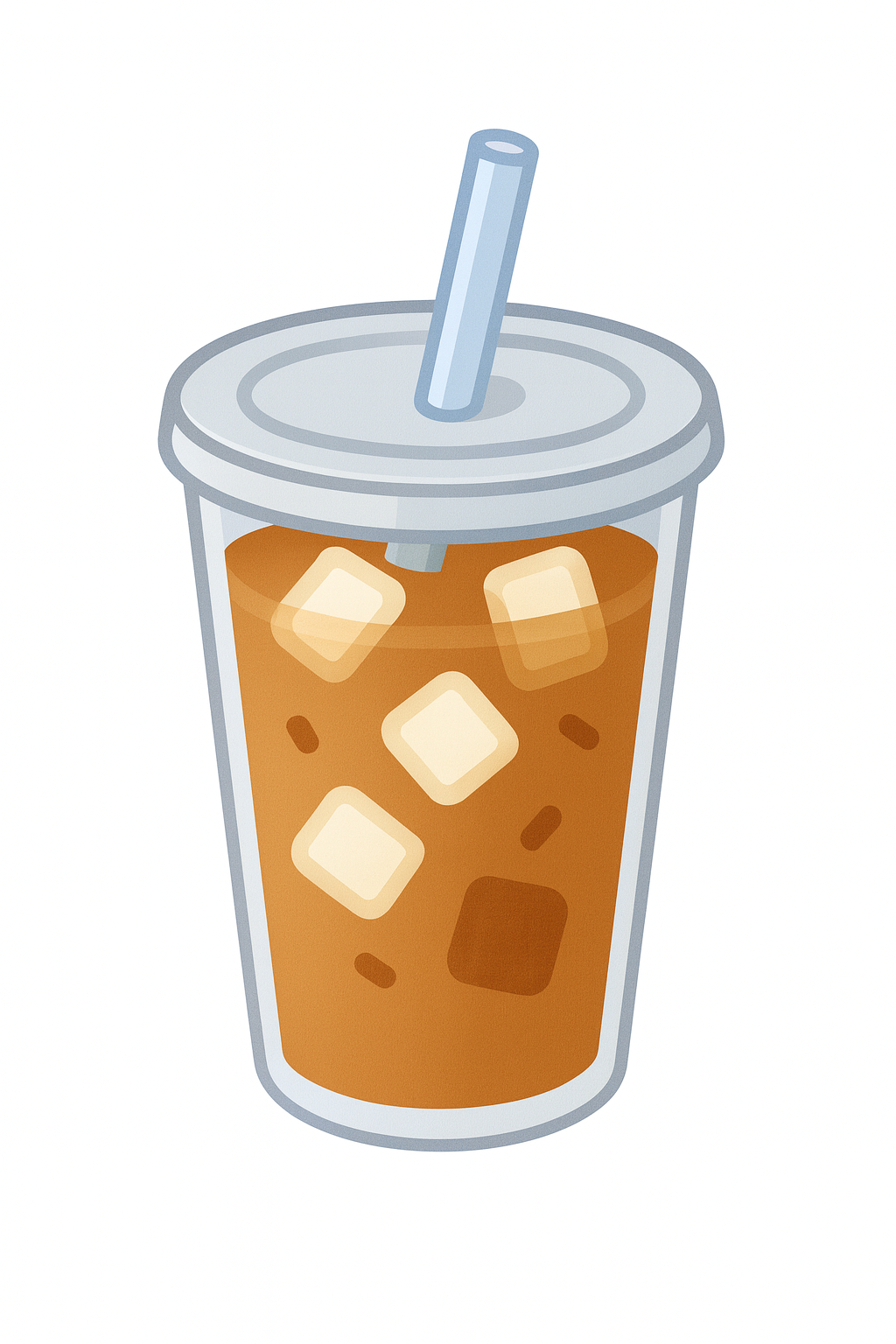}
\end{minipage}
\hspace{-0.4cm}
\begin{minipage}{1\textwidth}
    LATTE: Latent Trajectory Embedding \\for Diffusion-Generated Image Detection
\end{minipage}
}

\author{
Ana Vasilcoiu\thanks{Shared authorship. Correspondence to ana-maria.vasilcoiu@student.uva.nl and i.najdenkoska@uva.nl.}\hspace{2pt} \textsuperscript{1},
Ivona Najdenkoska\footnotemark[1]\hspace{4pt}\textsuperscript{1,2},
Zeno Geradts\textsuperscript{1,2},
Marcel Worring\textsuperscript{1} \\
\textsuperscript{1}University of Amsterdam, \textsuperscript{2}Netherlands Forensic Institute (NFI) \\
}

\iclrfinalcopy 
\begin{document}
\maketitle

\begin{abstract}
    The rapid advancement of diffusion-based image generators has made it increasingly difficult to distinguish generated from real images. This erodes trust in digital media, making it critical to develop generated image detectors that remain reliable across different generators. While recent approaches leverage diffusion denoising cues, they typically rely on single-step reconstruction errors and overlook the sequential nature of the denoising process. In this work, we propose \textbf{LATTE} - \textbf{Lat}ent \textbf{T}rajectory \textbf{E}mbedding - a novel approach that models the evolution of latent embeddings across multiple denoising steps. 
    Instead of treating each denoising step in isolation, LATTE captures the trajectory of these representations, revealing subtle and discriminative patterns that distinguish real from generated images.
    Experiments on several benchmarks, such as GenImage, Chameleon, and Diffusion Forensics, show that LATTE achieves superior performance, especially in challenging cross-generator and cross-dataset scenarios, highlighting the potential of latent trajectory modeling. The code is available on the following link: \url{https://github.com/AnaMVasilcoiu/LATTE-Diffusion-Detector}.
\end{abstract}
 
\section{Introduction}
Diffusion-based generative models have fundamentally transformed the field of image generation \citep{ddpm,ddim,ldm,glide,adm,imagen,sdxl,midjourney,flux}. These models generate photorealistic content - such as portraits, landscapes, and complex scenes - by iteratively adding and then removing noise from data or latent representations, typically guided by a text prompt \citep{sdv}. While this progress has unlocked transformative and creative applications, it has also facilitated the creation of fake images that are hard to visually distinguish from authentic content. Such capabilities have already been exploited by malicious actors, for instance, to create fraudulent impersonations of public figures \citep{twomey2023deepfake, de2023unethical} or fabricate ``evidence'' in legal disputes \citep{delfino2022deepfakes, sandoval2024threat, KoutrasSelvadurai2024}. The challenge is also amplified by the growing landscape of image generation models, each introducing its own artifacts and characteristics. This underscores the urgent need for robust detectors able to distinguish real from generated images.

Recent efforts to detect generated images \citep{dire,dnf,sedid,lare,ricker2024aeroblade,drct,fire,aide,cospy} leverage distinctive signatures left by the generative process. Based on the hypothesis that diffusion models can reconstruct synthetic images more accurately than real ones, methods like DIRE \citep{dire} and LaRE \citep{lare} define novel representations that capture the error between an input image and its reconstruction. 
While achieving solid performance, these approaches rely on single-step representations and overlook the inherent sequential nature of denoising - a process that largely underlies the synthetic artifacts of fake images. We address this by treating the sequence of latent representations as a distinctive signature.

In this paper, we introduce \textbf{Lat}ent \textbf{T}rajectory \textbf{E}mbedding - \textbf{LATTE}, a novel approach that explicitly models the evolution of latent representations across multiple denoising steps.
Namely, diffusion models generate images through a sequence of gradual denoising steps, where each learned update iteratively refines the sample toward the data manifold.
This iterative process defines a trajectory that reflects how the model interprets and refines the underlying content.
We hypothesize that real images, whose details and textures can lie outside the model’s learned manifold, will often produce small inconsistencies between successive denoising steps. On the contrary, fake images will follow smoother, more self-consistent trajectories aligned with the model's generative prior.
Specifically, given an image, we leverage a pretrained latent diffusion model to obtain its latent embedding. We apply standard forward noising and then extract intermediate latent states during the denoising at evenly spaced steps. This spacing provides a representative view of early, middle, and late denoising stages, capturing the full spectrum of the denoising dynamics. The resulting trajectory reflects how the internal representation evolves across steps, but it does not reveal which image regions drive these changes. To enrich the trajectory signal, we fuse each latent with visual features extracted from a pretrained image encoder using a stack of transformer decoders. The enriched sequence is subsequently aggregated into a compact representation, combined with global image features, and passed to a lightweight classifier. This combination of latent dynamics and semantic cues enables LATTE to leverage subtle inconsistencies indicative of generated content.

We evaluate LATTE on well-established benchmarks for generated image detection, namely GenImage \citep{genimage}, Chameleon \citep{aide}, and Diffusion Forensics \citep{dire}. Our model surpasses current state-of-the-art methods, achieving an average improvement of 4.1\% on GenImage over AIDE \citep{aide} and 7.1\% in cross-domain settings on Diffusion Forensics over LaRE \citep{lare}. In particular, on one of the most challenging subsets of GenImage i.e., BigGAN \citep{biggan}, LATTE outperforms the most competitive baseline by 9.5\%, highlighting its cross-generator generalizability. In cross-domain settings - for instance, the Bedroom partition of Diffusion Forensics - we observe a 11.1\% gain, underscoring LATTE’s robustness to specialized domains.

In summary, our contributions are threefold: (1) We propose LATTE, the first diffusion-based embedding that explicitly leverages the trajectory of latent states across multiple denoising steps. (2) We introduce a two-stage architecture that (i) samples and enriches latent trajectories via transformer decoders and (ii) aggregates the latent embeddings into a compact and discriminative representation. (3) We demonstrate that LATTE achieves state-of-the-art performance and exhibits strong performance across diverse benchmarks, unseen generators, perturbations, and domains.

\section{Related Work}

\noindent\textbf{Image Generation Models.}
Early methods for image generation were predominantly based on Generative Adversarial Networks (GANs) \citep{gan, karras2017progressive, biggan, stargan, spade, zhu2017unpaired}, Variational Autoencoders (VAEs) \citep{vae, sohn2015learning, infovae, vandenoord}, and autoregressive models \citep{pixelcnn, imagetransformer, taming, zeroshot}. GANs produce realistic images, but are hard to train and lack stable likelihood estimation. VAEs enable efficient inference and structured latent spaces but tend to generate blurry images. Autoregressive models offer precise likelihood modeling but suffer from slow, sequential sampling, especially at high resolutions.

To address the limitations of early methods, Denoising Diffusion Probabilistic Models (DDPMs) \citep{ddpm} introduced a generative process that reverses a gradual noising procedure, offering stable likelihood-based training and state-of-the-art image quality. 
Further advancements have explored improved sampling efficiency \citep{ddim}, accelerated solvers \citep{dpm_solver}, architectural refinements \citep{imagen, glide}, and improved conditional generation with classifier-free guidance \citep{cfguidance}. Latent Diffusion Models (LDMs) \citep{ldm} improved scalability by operating in a compressed latent space learned via a variational autoencoder, enabling high-resolution generation at much lower cost. LDMs underpin popular models like Stable Diffusion, and have enabled extensions such as ControlNet \citep{controlnet} for spatial conditioning, SDXL \citep{sdxl} for ultra-high-resolution output, and LCM \citep{lcm} for efficient few-step sampling.
Diffusion models now represent the primary focus of current research in generated image detection, as also addressed in this paper.

\noindent\textbf{Detection of Generated Images.}
Early efforts in generated image detection targeted GAN-generated content, starting with handcrafted features \citep{headpose, eyeblink} and later advancing to convolutional neural networks (CNNs) trained on datasets like FaceForensics++ \citep{faceforensics}. Subsequent works investigated intrinsic manipulation traces such as spectral artifacts in the frequency domain \citep{luo2021generalizing, frank2020leveraging} and inconsistencies in noise distributions \citep{noise1, noise2}. While these approaches improved robustness across GANs, they demonstrated limited generalization to diffusion-generated images.

To improve the generalizability of methods for detecting diffusion-based images, recent work has explored strategies that leverage the internal mechanics of the diffusion process. Some approaches focus on full image reconstruction: DIRE \citep{dire} introduced the idea of using DDIM \citep{ddim} inversion error as a discriminative feature, while DRCT \citep{drct} uses a contrastive training objective on reconstructed images. Other methods, like LaRE \citep{lare}, improve efficiency by operating in latent space and using a single-step inversion. AIDE \citep{aide} incorporates low-level patch statistics and high-level semantics. In contrast, our method leverages the trajectory of latent states across denoising steps, capturing the evolution of the process as a more discriminative representation.

Another line of research explores powerful vision encoders, such as CLIP \citep{radford2021learning}, used either as a frozen feature extractor with downstream classifiers \citep{mfclip}, or in fine-tuned multi-modal frameworks aligning image and text embeddings to capture inconsistencies in generated content \citep{raisingthebar, noiseassisted}. 
We also employ CLIP’s vision encoder, alongside other large-scale vision encoders, to enrich our proposed latent trajectory embedding.

\section{Methodology}
\label{sec:method}
In this section, we introduce our \textbf{Lat}ent \textbf{T}rajectory \textbf{E}mbedding (\textbf{LATTE}) for generated image detection. 
First, we give a brief overview of the denoising process in latent diffusion models. Then, we continue by introducing LATTE and explaining how to extract and fuse a sequence of latents with visual features. Finally, we show how LATTE can be aggregated into a unified representation to enhance the detection of generated images. 

\subsection{Preliminaries}
\label{sec:preliminaries}
\paragraph{Diffusion Probabilistic Models.}  Diffusion models define a Markov chain of diffusion steps that progressively add Gaussian noise to data until turning it into noise.
In the literature, this is referred to as a forward noising process \citep{ddpm}. Specifically, starting from a clean image $x$, the forward chain gradually injects Gaussian noise over \(T\) discrete steps:
\begin{equation}
    q(x_t \!\mid\! x_{t-1})
\;=\;
\mathcal{N}\!\bigl(x_t;\,\sqrt{\alpha_t}\,x_{t-1},\,(1-\alpha_t)\mathbf{I}\bigr),
\end{equation}
where $x_t$ is the noisy image at step $t$ and the schedule \(\{\alpha_t\}\) controls the noise variance at each step. After \(T\) steps, the image becomes nearly isotropic noise. In the reverse process, also defined as a Markov chain, the noisy image is gradually denoised to obtain the raw image. This backward chain leverages a neural network \(\epsilon_\theta(x_t,t)\) parameterized by $\theta$ to predict and remove this noise, defined as:
\begin{equation}
    p_\theta(x_{t-1}\!\mid\!x_t)
=\mathcal{N}\!\bigl(x_{t-1};\,\tfrac{1}{\sqrt{\alpha_t}}\bigl(x_t - (1-\alpha_t)\epsilon_\theta(x_t,t)\bigr),\,\sigma_t^2\mathbf{I}\bigr).
\end{equation}

\paragraph{Latent Diffusion.}  
To improve efficiency, latent diffusion models \citep{ldm} first encode images into a lower-dimensional latent space via a pretrained VAE encoder \(E_{\mathrm{VAE}}\), producing \(z_0 = E_{\mathrm{VAE}}(x_0)\). The forward and reverse processes then operate on these latent codes \(z_t\in\mathbb{R}^d\):
\begin{align}
\label{eq:num-3}
    q(z_t \!\mid\! z_{t-1})
    &= \mathcal{N}\!\bigl(z_t;\,\sqrt{\alpha_t}\,z_{t-1},\,(1-\alpha_t)\mathbf{I}\bigr), \\
    \label{eq:num-4}
    p_\theta(z_{t-1}\!\mid\! z_t)
    &= \mathcal{N}\!\bigl(z_{t-1};\,\mu_\theta(z_t,t),\,\Sigma_\theta(z_t,t)\bigr).
\end{align}
After denoising to \(z_0\), a VAE decoder \(D_{\mathrm{VAE}}\) reconstructs the final image \(\hat x = D_{\mathrm{VAE}}(z_0)\). Latent diffusion thus preserves high sample quality while reducing computational and memory demands.

\begin{figure}[t]
    \centering
    \includegraphics[width=\textwidth]{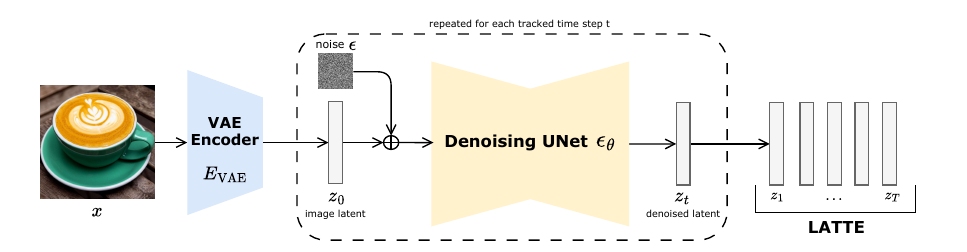}
    \caption{\textbf{Extraction of LATTE representation.} We construct the LATTE sequence by performing a single-step reconstruction for a selection of timesteps throughout the whole trajectory.
    }
    \label{fig:latte_extraction}
\end{figure}

\subsection{LATTE: Latent Trajectory Embedding}
In diffusion models, an image is reconstructed from noise by iteratively denoising latent variables over a sequence of timesteps (see Eqs.~(\ref{eq:num-3})–(\ref{eq:num-4})).
LATTE leverages the sequential structure of diffusion models by explicitly modeling how the latent embedding evolves across denoising steps. Instead of performing the full reverse chain, which is computationally expensive, we approximate intermediate states using single-step denoising at selected timesteps.

Given an input image \(x\), we first encode it into latent space using a pretrained VAE encoder: $z_0 = E_{\mathrm{VAE}}(x)$, as explained in section \ref{sec:preliminaries}. Next, for each selected timestep \(t\), we simulate the forward diffusion process in one closed‐form operation, producing a noisy latent:
\begin{equation}
    z_{t} = \sqrt{\bar\alpha_{t}}\,z_0 + \sqrt{1 - \bar\alpha_{t}}\,\epsilon,
\quad \epsilon \sim \mathcal{N}(0, I),
\end{equation}
where \(\bar\alpha_{t} = \prod_{s=1}^{T} \alpha_s\) accumulates the noise schedule up to \(t\). We then approximate the reverse diffusion at \(t\) by performing a single denoising update using the pretrained UNet’s noise predictor \(\epsilon_\theta\):
\begin{equation}
    \hat z_{t} = z_{t} - \sqrt{1 - \alpha_{t}}\,\epsilon_\theta(z_{t}, t).
\end{equation}

This one‐step correction yields an estimate \(\hat z_{t}\) that closely approximates the latent at \(t\), while avoiding the overhead of a full reverse pass from \(T\) to \(t\). By repeating this forward‐then‐single‐step reverse procedure for each of the $T$ timesteps \(\{t, \ldots, T\}\) chosen to uniformly span the denoising schedule, we assemble the latent trajectory embeddings: $\mathcal{T}(x) = \{\hat z_{1}, \hat z_{2}, \dots, \hat z_{T}\}$, illustrated in Figure \ref{fig:latte_extraction}. 

\subsection{Architecture Details}
\begin{figure}[t]
    \centering
    \includegraphics[width=\textwidth]{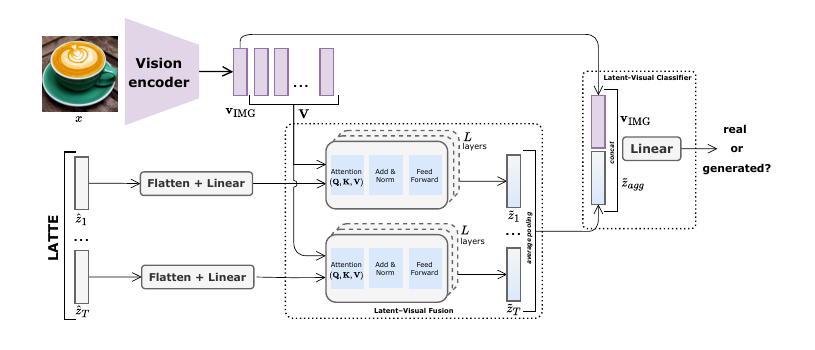}
    \vspace{-0.8cm}
    \caption{\textbf{Overview of our proposed architecture using LATTE.} It encompasses two stages: (1) \textit{Latent–Visual Fusion}, where the LATTE is fused with visual semantics through stacks of $L$ cross-attention layers, and (2) \textit{Latent-Visual Classifier} for average aggregation and output prediction. }
    \label{fig:latte_architecture}
\end{figure}

Our architecture grounds the latent trajectory in visual context, ensuring that the latent representations remain tied to the image content. As illustrated in Figure \ref{fig:latte_architecture}, it unifies two complementary feature streams - the LATTE sequence and visual semantics - through two main stages: \textit{Latent–Visual Fusion} and \textit{Latent-Visual Classifier}.

\noindent\textbf{Latent–Visual Fusion.} 
Each latent embedding is enhanced through cross-attention with spatial features extracted from a pretrained vision encoder, to ground the denoising trajectory in the image content.
Given an input image $x$, the vision encoder produces two outputs: (1) patch-level visual embeddings $\mathrm{V} \in \mathbb{R}^{N \times d}$, and (2) a global image token $\mathbf{v}_{\text{IMG}} \in \mathbb{R}^d$. The patch tokens $\mathrm{V}$ capture fine-grained spatial and semantic information from the image and are leveraged for the refinement of the LATTE representation. The $\mathbf{v}_{\text{IMG}}$ token provides a holistic representation of the image and is used in the second stage. 
 
Each latent embedding $\hat z_{t}$ in the trajectory $\mathcal{T}(x)$ is first flattened and linearly projected to match the dimensions $d$ of the visual features $\mathrm{V}$. The projected latents are then independently enhanced using a stack of $L$ transformer decoders, each consisting of a cross-attention layer followed by a feed-forward layer, with residual connections and layer normalization. Specifically, each latent $\hat z_{t}$ attends to the patch-level visual embeddings $\mathrm{V}$ using multi-head cross-attention (MHA) mechanism:
\vspace{-0.1cm}
\begin{equation}
    \tilde z_{t} = \mathrm{MHA}(Q, K, V) = [{head}_1, \dots {head}_h]\mathbf{W}^O, \mathrm{where} \: {head}_i =  \text{softmax} \left( \frac{\hat z_{t} K^\top}{\sqrt{d}} \right) V,
\end{equation}
where the keys and values $K, V \in \mathbb{R}^{N \times d}$ are both the visual features $\mathrm{V}$, $\mathbf{W}^O$ is the output projection layer, $h$ is the number of heads and $d$ is the dimension of the embeddings. Each $\hat z_{t}$ is processed through $L$ such layers, allowing it to align with different spatial features in the image independently of the other timesteps.

\noindent\textbf{Latent-Visual Classifier.}
After enhancing each latent embedding through $L$ transformer decoder layers, we obtain a set of enriched embeddings ${\tilde {z}_1, \ldots, \tilde {z}_T}$. To aggregate this sequence into a unified representation, we perform average pooling across all $T$ latents: $\tilde z_{\text{agg}} = \frac{1}{T} \sum_{i=1}^{T} \tilde {z}_t$. Alternatively, we can perform $CLS$ token pooling where a special token ${z}_{\text{CLS}}$ is prepended to the sequence of latents ${{z}_{\text{CLS}}, \tilde {z}_1, \ldots, \tilde {z}_T}$, processed with self-attention layers and then used as an aggregation $\tilde z_{\text{agg}}$.
The aggregated representation encodes how the latent embeddings transition through successive denoising steps, effectively encoding the reconstruction trajectory. Next, to incorporate a holistic semantic-level context, we concatenate $\tilde z_{\text{agg}}$ with the global image token $\mathbf{v}_{\text{IMG}}$: $\mathbf{z} = [\tilde z_{\text{agg}} \mathbin\Vert \mathbf{v}_{\text{IMG}}] \in \mathbb{R}^{2d}$.

Finally, we feed this joint embedding $\mathbf{z}$ into a lightweight linear classifier, which leverages this combined information to make a real-vs-generated prediction. By pooling the latent embeddings and grounding them in image semantics, our aggregation strategy effectively amplifies subtle artifacts that single-step or pixel-based methods tend to overlook.

\section{Experiments \& Results}
\subsection{Experimental Setup}
\label{sec:experimental_setup}

\textbf{Datasets.} 
We evaluate LATTE across several complementary settings.
First, to assess overall detection and cross-generator robustness, we use the GenImage dataset\footnote{\scriptsize Licensed under CC BY-NC-SA 4.0.}\citep{genimage}, which contains real and fake images generated by eight generative models, including diffusion- and GAN-based approaches. Next, to test performance under more visually-challenging scenarios, we use the Chameleon dataset \citep{aide}, which includes high-quality synthetic images designed to reduce detection artifacts.
To evaluate cross-domain generalization, we use the Diffusion Forensics dataset \citep{dire}, which spans multiple visual domains such as bedrooms, churches, and faces. All images are resized to $224 \times 224$ before being passed to the model. 

\textbf{Training \& Evaluation.} We extract the latent trajectories using Stable Diffusion 2.1. We empirically chose five timesteps: [981, 741, 521, 261, 1] for extracting the latents, which evenly spread across the trajectory. The visual features are obtained with a pretrained ConvNeXt \citep{convnext}, yielding a dimension size of $512$. All models are trained by minimizing binary cross-entropy loss to convergence, monitored on a held-out validation split matching the training generator. We use a batch size of 32, AdamW \citep{loshchilov2017decoupled} optimizer (lr = 1e-4, weight decay = 4e-5), and a cosine-annealed learning rate scheduler.
The experiments are conducted on a single H100 GPU, by training for approximately 2 hours for a single epoch. 
To provide a comprehensive evaluation, we follow standard practice in detection tasks and evaluate our models using accuracy and average precision. 
The code repository, training, and evaluation details will be released.

\subsection{Comparison to Baselines}
We first evaluate our method on GenImage \citep{genimage}, which essentially tests cross-generator generalization. All models are trained on images generated by SDv1.4 and evaluated across eight different generators, with baseline results cited from \cite{aide}. The results, shown in Table \ref {tab:acc_comparison}, indicate that LATTE/\texttt{Avg} (using average pooling) achieves the highest average accuracy among a broad set of related methods, improving by 4.1\% over the recent AIDE model, followed by LATTE/\texttt{CLS} (using CLS token pooling). Notably, on one of the most challenging subsets - BigGAN, LATTE/\texttt{Avg} surpasses the strongest prior baseline \citep{univfd} by $9.5$\%. Note that we continue using LATTE/\texttt{Avg} in the subsequent experiments as our best model, denoted as LATTE for brevity.

\begin{table}[h]
\centering
\caption{\textbf{Comparison of LATTE to baselines on GenImage benchmark \citep{genimage}.} All methods are trained on SDv1.4 of GenImage and evaluated over eight image generators. LATTE/\texttt{Avg} achieves the best average accuracy, improving by $4.1\%$ over state-of-the-art methods. }
\label{tab:acc_comparison}
\vspace{-0.2cm}
\resizebox{\textwidth}{!}{
\begin{tabular}{lcccccccc|c}
\toprule
\textbf{Method} & Midjourney & SD v1.4 & SD v1.5 & ADM & GLIDE & Wukong & VQDM & BigGAN & \textit{\textbf{Avg.}} \\
\midrule
CNNSpot \citep{wang2020cnn}     & 52.8 & 96.3 & 95.9 & 50.1 & 39.8 & 78.6 & 53.4 & 46.8 & 64.2 \\
F3Net \citep{f3net}       & 50.1 & 99.9 & \textit{99.9} & 49.9 & 50.0 & 99.9 & 49.9 & 49.9 & 68.7 \\
Spec \citep{zhang2019detecting}  & 52.0 & 99.4 & 99.2 & 49.7 & 49.8 & 94.8 & 55.6 & 49.8 & 68.8 \\
GramNet \citep{gramnet}   & 54.2 & 99.2 & 99.1 & 50.3 & 54.6 & 98.9 & 50.8 & 51.7 & 69.9 \\
DeiT-S \citep{deits}  & 55.6 & 99.9 & 99.8 & 49.8 & 58.1 & 98.9 & 56.9 & 53.5 & 71.6 \\
ResNet-50 \citep{rn50} & 54.9 & 99.9 & 99.7 & 53.5 & 61.9 & 98.2 & 56.6 & 52.0 & 72.1 \\
DIRE \citep{dire}    & 65.8 & 99.7 & 99.7 & 54.5 & 58.1 & 99.4 & 54.3 & 49.8 & 72.7 \\
UnivFD \citep{univfd}    & 73.2 & 84.2 & 84.0 & 55.2 & 76.9 & 75.6 & 56.9 & 80.3 & 73.3 \\
Swin-T \citep{swint}       & 62.1 & 99.9 & 99.8 & 49.8 & 67.6 & 99.1 & 62.3 & 57.6 & 74.8 \\
GenDet \citep{zhu2023gendet}  & 89.6 & 96.1 & 96.1 & 58.0 & 78.4 & 92.8 & 66.5 & 75.0 & 81.6 \\
DRCT \citep{drct} & \textbf{94.6} & 99.8 & 99.8 & 61.8 & 65.9 & 99.9 & 74.8 & 58.8 & 82.1 \\
PatchCraft \citep{patchcraft} & 79.0 & 89.5 & 89.3 & 77.3 & 78.4 & 89.3 & 83.7 & 72.4 & 82.3 \\
Co-Spy \citep{cospy}   & 83.4 & 96.8 & 96.7 & 67.2 & 93.0 & 95.9 & 78.8 & 65.2 & 84.6 \\
LaRE \citep{lare} & 74.0 & 100 & 99.9 & 61.7 & 88.5 & \textbf{100} & \textbf{97.2} & 68.7 & 86.2 \\
AIDE \citep{aide}    & 79.4 & 99.7 & 99.8 & 78.5 & 91.8 & 98.7 & 80.3 & 66.9 & 86.9 \\
\midrule
\textbf{LATTE}/\texttt{CLS} & 85.8 & 99.0 & 99.0 & \textbf{83.2} & 86.7 & 96.8 & 88.7 & 77.8 & 89.6 \\
\rowcolor{yellow!15}
\textbf{LATTE}/\texttt{Avg} & 88.8 & \textbf{100} &  \textbf{99.9} & 74.0 & \textbf{95.8} & 98.9 & 80.8 & \textbf{89.8} & \textbf{91.0} \\ 
\bottomrule
\end{tabular}
}
\end{table}

\begin{figure}[h]
    \centering
    \includegraphics[width=\textwidth]{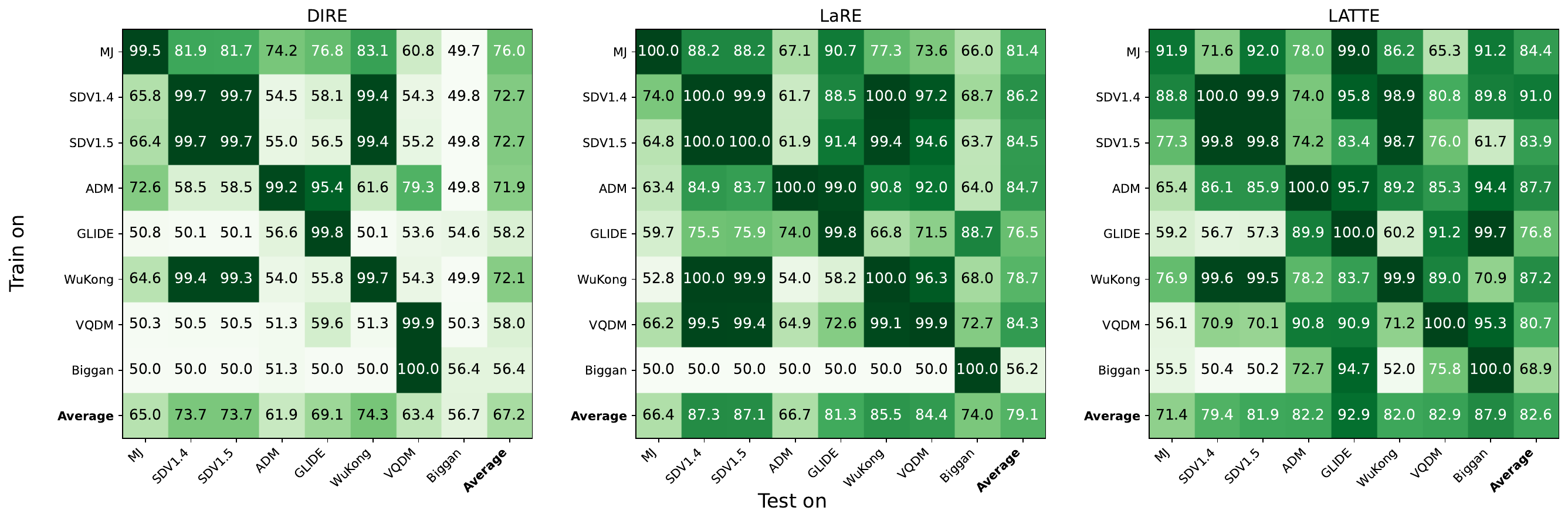}
    \vspace{-0.6cm}
    \caption{\textbf{Comparison of LATTE to baselines, by training and testing across all 8 generators of GenImage.} Each plot corresponds to one detector - DIRE (left; baseline), LaRE (center; baseline), and LATTE (right; proposed) - and shows the accuracy(\%) when training on the subset listed on the vertical axis and testing on the subset listed along the horizontal axis.}
    \label{fig:acc_comparison_greens}
\end{figure}

Next, we train our model on each generator-specific subset of GenImage and test it across all other subsets. Figure \ref{fig:acc_comparison_greens} reports the averaged accuracies, where our model again achieves the best overall performance.
These findings demonstrate that explicitly modeling the trajectory evolution in latent space yields stronger robustness and more reliable detection across diverse image generators. 


We further evaluate our model on Chameleon \citep{aide}, a recently proposed benchmark designed to reflect real-world scenarios by covering a broad range of content, including humans, animals, objects, and scenes. This benchmark allows us to test how well the model generalizes beyond its training distribution and captures transferable representations. As shown in Table \ref{tab:chameleon}, our method achieves consistent improvements over the baselines, achieving 2.5\% improvement over AIDE when trained on the GenImage dataset. The results highlight both the robustness of our approach and its effectiveness in generalizing across diverse visual domains.

Finally, we evaluate the cross-domain generalization ability of LATTE on the Diffusion Forensics dataset \citep{dire}. Specifically, we train LATTE on the SDv1.4 subset of GenImage and use LaRE and AIDE trained on the same data for a fair comparison. Table \ref{tab:model-comparison-split} reports accuracies across various generators and three distinct domains - Bedroom, ImageNet, and CelebA - which differ substantially from GenImage in both content and style. Across all three domains, LATTE consistently outperforms both LaRE and AIDE, achieving improvements such as 11.1\% on Bedroom and 4\% on Imagenet, with an overall average gain of 7.1\%. We also notice that in-domain performance (train and test on the same data) is already saturated in prior work - often reported at or near 100\% - so it offers limited insight into generalization.
Therefore, we prioritized the evaluation in cross-generator and cross-domain settings.

\begin{table}[t]
\centering
\caption{\textbf{Cross-domain comparison on Chameleon \citep{aide}.} Each column represents the accuracy (\%) of different detectors, and the rows indicate the used training set. 
}
\vspace{-0.25cm}
\label{tab:chameleon}
\begin{adjustbox}{width=\textwidth}
\begin{tabular}{l|ccccccccc|c}
\toprule
 \textbf{Training set} & FreDect & LNP & UnivFD & DIRE & NPR & PatchCraft & CNNSpot & GramNet & AIDE & \textbf{LATTE} \\
\midrule
SDv1.4 & 56.9 & 58.5 & 55.6 & 59.7 & 58.1 & 56.3 & 60.1 & 60.9 & 62.6 & \cellcolor{yellow!15}\textbf{63.8} \\
All GenImage & 57.2 & 58.5 & 60.4  & 57.8  & 57.8  & 55.7 & 60.9   & 59.8  & 65.8 & \cellcolor{yellow!15} \textbf{68.3} \\
\bottomrule
\end{tabular}
\end{adjustbox} 

\end{table}

\begin{table}[htbp]
    \centering
    \caption{\textbf{Cross-domain comparison on Diffusion Forensics \citep{dire}}. LATTE achieves an overall average improvement of 7.1\% accuracy over LaRE and 14.8\% over AIDE.}
    \vspace{-0.2cm}    
    \subfloat[\textbf{Bedroom}]{
        \begin{adjustbox}{width=\textwidth}
            \begin{tabular}{l|cccccccccccc|cc}
                \toprule
                \textbf{Method} & ADM & Dalle2 & DDPM & ProjGAN & StyleGAN & IDDPM & IF & LDM & MidJ & SDV & PNDM & VQDM & \textit{\textbf{Avg.}} \\
                \midrule
                LaRE  & 53.0 & 66.7 & 57.6 & 50.5 & 62.3 & 55.2 & 90.9 & 93.5 & 90.9 & 78.4 & 53.5 & 82.4 & 69.5 \\
                AIDE  & 66.5 & \textbf{77.9} & \textbf{66.5} & 55.5 & 76.0 & 57.0 & 94.6 & 79.0 & \textbf{94.9} & 88.4 & 54.7 & 84.2 & 74.6 \\
                \rowcolor{yellow!15}
                \textbf{LATTE} & \textbf{89.9} & 76.5 & 63.3 & \textbf{65.3} & \textbf{93.5} & \textbf{90.0} & \textbf{99.9} & \textbf{99.3} & 91.2 & \textbf{91.9} & \textbf{75.3} & \textbf{92.6} & \textbf{85.7} \\
                \bottomrule
            \end{tabular}
        \end{adjustbox}
    }
    \vspace{-0.5em}
    \noindent
    \begin{minipage}[t]{0.49\textwidth}
        \centering
        \subfloat[\textbf{ImageNet}]{
            \begin{adjustbox}{width=0.75\textwidth}
                \begin{tabularx}{\linewidth}{lXX|X}
                    \toprule
                    \textbf{Method} & ADM & SDV & \textit{\textbf{Avg.}} \\
                    \midrule
                    LaRE  & 81.4 & 98.5 & 89.9 \\
                    AIDE & 53.6 & \textbf{98.9} & 76.2 \\
                    \rowcolor{yellow!15}
                    \textbf{LATTE} & \textbf{89.8} & 98.0 & \textbf{93.9} \\
                    \bottomrule
                \end{tabularx}
            \end{adjustbox}
        }
    \end{minipage}
    \hfill
    \begin{minipage}[t]{0.49\textwidth}
        \centering
            \subfloat[\textbf{Celeba}]{
            \begin{adjustbox}{width=0.75\textwidth}
                \begin{tabularx}{\linewidth}{lXXXX|X}
                    \toprule
                    \textbf{Method} & Dalle2 & IF & MidJ & SDV & \textit{\textbf{Avg.}} \\
                    \midrule
                    LaRE  & \textbf{77.7} & 96.3 & \textbf{90.9} & 95.2 & 90.0 \\
                    AIDE & 70.8 & 76.5 & 69.5 & 85.2 & 75.5 \\
                    \rowcolor{yellow!15}
                    \textbf{LATTE} & 77.5 & \textbf{96.7} & \textbf{90.9} & \textbf{99.6} & \textbf{91.1} \\
                    \bottomrule
                \end{tabularx}
            \end{adjustbox}
        }
    \end{minipage}
    \label{tab:model-comparison-split}
\end{table}
\vspace{-0.4cm}

\subsection{Ablation Study}
\label{sec:ablation}
In this section, we present ablation studies to quantify the contribution of each component, the impact of the denoising steps and vision backbone. Additional ablations are available in the Appendix \ref{appendix:ablations}.

\noindent\textbf{Importance of each component.}
We conduct an ablation study on three components: the visual features from the backbone, the latent trajectory from intermediate diffusion steps, and the Latent–Visual Fusion module that aligns these modalities via cross-attention. Four model variants are evaluated: (A) visual features only, (B) latent trajectory only, (C) visual + latent trajectory without fusion, and (D) the full model with all components. As shown in Table \ref{tab:ablation-full}, both visual-only (A) and latent-only (B) variants perform poorly, confirming that neither modality alone is sufficient. Combining the two in (C) improves performance, indicating complementary cues, but the gains remain limited. The full model (D) achieves the best results across nearly all subsets, with large improvements on challenging cases such as VQDM (+11.6\%) and BigGAN (+13.9\%), underscoring the importance of effectively fusing latent and visual information.

\begin{table}[h]
\centering
\caption{\textbf{Ablation on visual and latent components.} \ding{51} indicates that the component is included. Results are shown as Accuracy (\%) for each generator. Including all components of our approach outperforms the visual-only and latent-only configurations by 16.1\% and 37.8\%.}
\vspace{-0.3cm}
\label{tab:ablation-full}
\resizebox{\textwidth}{!}{%
\begin{tabular}{lccc|cccccccc|c}
\toprule
\textbf{Model} & \textbf{Visual} & \textbf{Latent} & \textbf{Fusion} 
& Midjourney & SDV1.4 & SDV1.5 & ADM & GLIDE & Wukong & VQDM & BigGAN & \textit{\textbf{Avg.}} \\
\midrule
A & \ding{51} & \ding{55} & \ding{55} & 83.5 & 99.9 & \textit{99.9} & 51.7 & 56.2 & \textbf{99.9} & 58.4 & 50.0 & 74.9 \\
B & \ding{55} & \ding{51} & \ding{55} & 50.0 & 58.3 & 58.4 & 50.0 & 50.0 & 56.6 & 52.6 & 50.0 & 53.2 \\
C & \ding{51} & \ding{51} & \ding{55} & 80.5 & \textit{100} & \textit{99.9} & \textbf{76.7} & 84.3 & 99.8 & 69.2 & 75.8 & 85.7 \\
\rowcolor{yellow!15}
D & \ding{51} & \ding{51} & \ding{51} & \textbf{88.7} & \textbf{100} & \textbf{99.9} & 74.0 & \textbf{95.7} & 98.9 & \textbf{80.8} & \textbf{89.7} & \textbf{91.0} \\
\bottomrule
\end{tabular}
}
\vspace{-0.3cm}
\end{table}

\paragraph{Influence of denoising steps.}
We study how performance changes with the number of denoising steps, varying $n \in \{1, 3, 5, 9, 13, 15\}$ used to sample intermediate latents for the trajectory. For $n=5$ steps, we empirically select the following: [981, 741, 521, 261, 1], while $n=1$ corresponds to the single midpoint t = 521. The remaining configurations include both endpoints ($t=1$ and $t \approx 1000$) with additional steps interpolated evenly across the trajectory. Our choice of such evenly spaced steps - spanning from near the start to the end of the trajectory - aims to capture the full spectrum of denoising behavior.
As shown in Table \ref{tab:choice_t}, accuracy improves as the number of sampled steps increases, peaking at $n = 5$. Beyond this point, the $n = 9$ configuration maintains competitive results, but performance declines at $13$ and $15$ steps, suggesting that adding more steps introduces redundancy rather than additional useful information.

\begin{table}[t]
\centering
\caption{\textbf{Accuracy(\%) comparison of varying lengths of latent trajectory.} We compare the effect of different timestep configurations on the average accuracy across eight generative models. The best accuracy is achieved with the 5-timestep configuration ($n = 5$).}
\vspace{-0.25cm}
\label{tab:choice_t}
\resizebox{\textwidth}{!}{
\begin{tabular}{c|cccccccc|c}
\toprule
\textbf{$n$}-steps & Midjourney & SDv1.4 & SDv1.5 & ADM & GLIDE & Wukong & VQDM & BigGAN & \textit{\textbf{Avg.}} \\
\toprule
1 & 80.4 & 99.2 & 98.8 & 71.2 & 85.6 & 96.9 & 74.2 & 62.5 & 83.6 \\
3 & 78.7 & 99.3 & 99.0 & \textbf{75.3} & 81.8 & 97.2 & 80.4 & 68.0 & 85.0 \\
\rowcolor{yellow!15}
\textbf{5} & \textbf{88.8} & \textbf{100} & \textbf{99.9} & 74.0 & \textbf{95.8} & \textbf{98.9} & \textbf{80.8} & \textbf{89.8} & \textbf{91.0} \\
9 & 86.2 & 99.5 & 99.4 & 75.2 & 94.4 & 98.0 & 79.8 & 86.9 & 89.9 \\
13 & 77.3 & 99.7 & 99.4 & 72.2 & 82.8 & 98.5 & 77.3 &	69.4 & 84.6 \\
15  &  77.3 &  99.6 &  99.3 &  73.5 &  82.9 &  98.4 &  78.3 & 62.7 &  84.0  \\
\bottomrule
\end{tabular}
}
\end{table} 

\textbf{Influence of vision backbone.}  In our preliminary experiments, we used CLIP encoders (RN50, ViT-B/32), which underperformed on GenImage. This prompted the shift to other backbones: ConvNeXt-Base \citep{convnext} pretrained on ImageNet-22k and CLIP ViT-L/14 \citep{ilharco_gabriel_2021_5143773}, also leveraged by \cite{univfd}. Both improved the results significantly, with ConvNeXt consistently achieving the highest accuracy, as demonstrated in Table \ref{tab:backbone}.

\begin{table}[h]
\centering
\caption{\textbf{Accuracy(\%) comparison between different vision backbones.} ConvNeXt outperforms CLIP ViT-L/14 by 5.3\%.}
\vspace{-0.25cm}
\label{tab:backbone}
\resizebox{\textwidth}{!}{
\begin{tabular}{l|cccccccc|c}
\toprule
\textbf{Vision encoder} & Midjourney & SDv1.4 & SDv1.5 & ADM & GLIDE & Wukong & VQDM & BigGAN & \textit{\textbf{Avg.}} \\
\toprule
CLIP ViT-L/14 & \textbf{98.2} & \textit{100} & \textbf{100} & 68.4 & 91.8 & \textbf{99.9} & 58.9 & 68.4 & 85.7 \\
\rowcolor{yellow!15}
\textbf{ConvNeXt} & 88.8 & \textbf{100} & 99.9 & \textbf{74.0} & \textbf{95.8} & 98.9 & \textbf{80.8} & \textbf{89.8} & \textbf{91.0} \\
\bottomrule
\end{tabular}
}
\end{table}

\vspace{-0.1cm}

\subsection{Embedding Space Analysis}
\begin{figure}[bh]
    \centering
    \includegraphics[width=\textwidth]{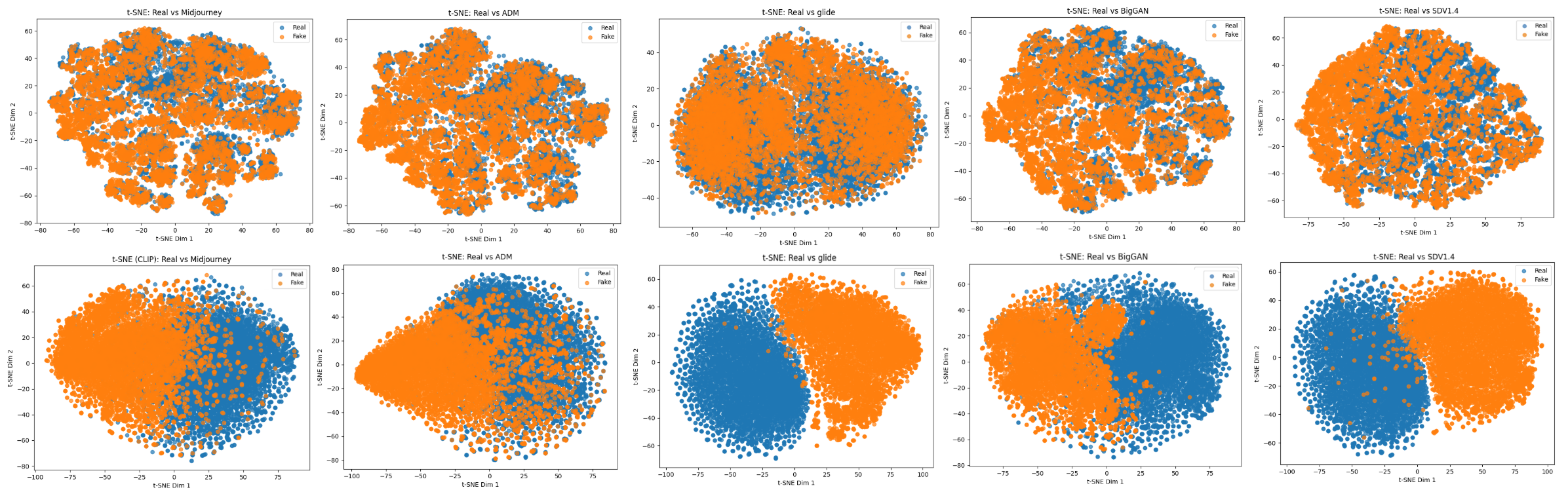}
    \vspace{-0.5cm}
    \caption{\textbf{Visualizations of t-SNE embeddings for real and fake images across five generators from GenImage.} The first row presents embeddings before using LATTE (extracted using the ConvNeXt), while the second row shows embeddings derived from LATTE. The much clearer separation in the second row illustrates LATTE's discriminative power.}
    \label{fig:embeddings}
\end{figure}

We assess our model’s discriminative capacity by visualizing real and generated image embeddings with t-SNE \citep{vandermaaten08a} in Figure \ref{fig:embeddings}.
The first row depicts the embeddings extracted from the original, frozen ConvNeXt backbone, while the second row displays embeddings after the backbone has been fine-tuned with LATTE. The embeddings in the second row exhibit much clearer separation between real (\textcolor{blue}{blue}) and generated (\textcolor{orange}{orange}) samples, indicating reduced overlap and stronger class separation. 

\subsection{Robustness to Unseen Perturbations}
\begin{figure}[h]
    \centering
    \includegraphics[width=\textwidth]{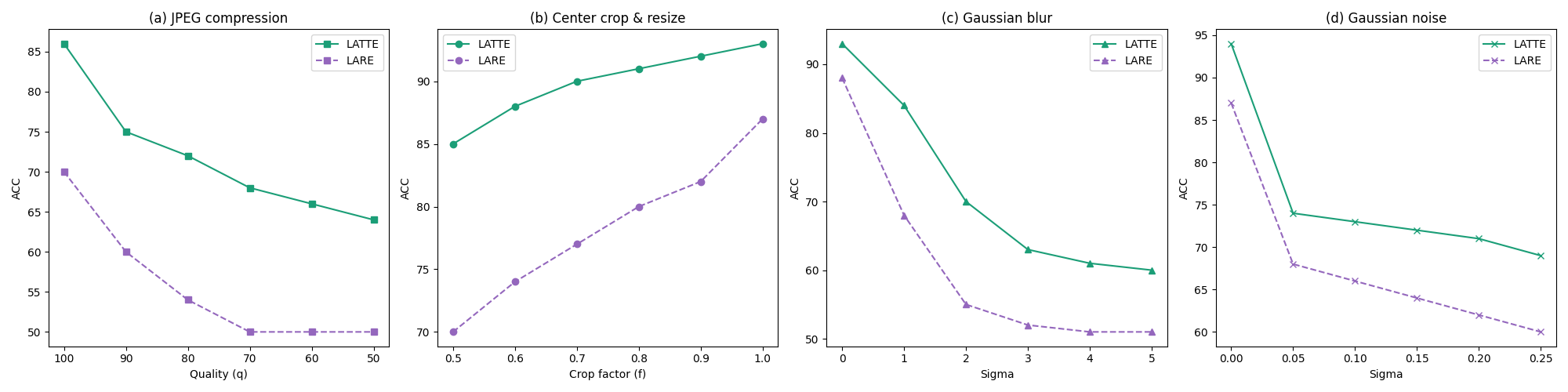}
    \vspace{-0.5cm}
    \caption{\textbf{Accuracy(\%) of LATTE vs. LaRE on perturbed images.} We evaluate and compare the robustness of both methods under four common transformations: JPEG compression, center crop \& resize, Gaussian blur, and noise. LATTE consistently outperforms LaRE across all perturbations.}
    \label{fig:robustness}
\end{figure}

We assess the robustness of LATTE under common post-processing operations like compression, resizing, Gaussian blur, and Gaussian noise. Such perturbations often occur in real-world pipelines and can severely degrade the subtle artifacts that detection methods depend on. As shown in Figure \ref{fig:robustness}, LATTE consistently outperforms LaRE, maintaining higher detection accuracy and greater stability. This shows that LATTE's reliance on multi-step latent trajectories is more invariant under such transformations than single-step reconstruction errors.

\vspace{-0.2cm}
\vspace{-0.15cm}

\subsection{Qualitative Analysis}
\begin{figure}[H]
    \centering
    \includegraphics[width=\textwidth]{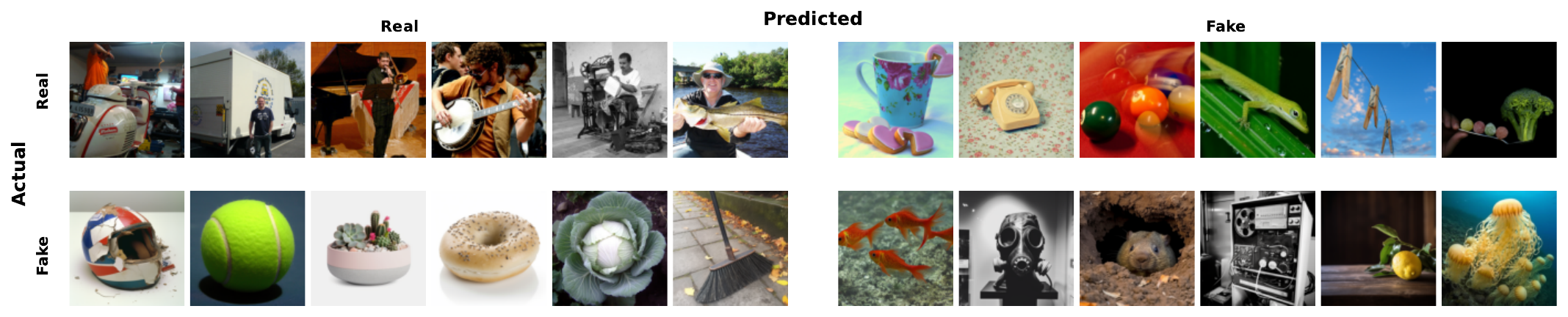}
    \vspace{-0.6cm}
    \caption{\textbf{Qualitative results in a confusion-matrix-style layout.} The rows show actual labels, and the columns show predictions of LATTE. }
    \label{fig:conf_matrix}
\end{figure}

We present qualitative examples in a confusion-matrix-style layout in Figure \ref{fig:conf_matrix}, highlighting representative model successes and failures.
\textbf{Top-left}: Real images with complex textures, human subjects, or fine structures are typically recognized as authentic, since such details remain difficult for generative models to replicate. \textbf{Top-right:} In contrast, some real images with smooth textures, saturated colors, or stylized lighting are misclassified as fake, reflecting the model’s sensitivity when authentic content visually resembles synthetic imagery. \textbf{Bottom-left:} On the other hand, high-quality generated images that appear simple or artifact-free may be mistaken for real, highlighting the difficulty of detecting visually convincing fakes. \textbf{Bottom-right:} Lastly, LATTE succeeds in correctly identifying visually convincing fake images, which suggests that it leverages subtle traces rather than only visual artifacts.

\vspace{-0.2cm}

\section{Conclusion}
We propose \textbf{LATTE}, a novel diffusion-generated image detection approach that models the sequential evolution of latents across multiple denoising steps. By capturing trajectory patterns and grounding them with visual features, LATTE learns a compact and discriminative representation. Experiments on GenImage, Chameleon, and Diffusion Forensics demonstrate that LATTE achieves state-of-the-art performance, including significant gains in cross-generator and cross-domain scenarios. Overall, this work highlights latent trajectory modeling as a new direction for generated image detection.

\textbf{Limitations.} While LATTE achieves strong performance and improved generalization, its effectiveness diminishes under strong post-processing (e.g., heavy JPEG compression or strong blur), indicating sensitivity to distribution shifts. Additionally, like most global detectors, LATTE has been evaluated primarily on fully synthetic versus real images, while detecting small, localized forgeries remains a distinct challenge for future work.

\section*{Ethics Statement} 
This work advances the field of synthetic media forensics by improving the detection of generated images. As generative models improve their ability to produce highly realistic content, frameworks or tools like LATTE, play an important role in combating disinformation, verifying content authenticity, and maintaining public trust in digital media. 

However, the deployment of the detection system also raises important ethical and societal considerations. As detection technologies improve, so do adversaries' strategies for evading them, potentially resulting in an arms race between generation and detection. Furthermore, there is a risk that such tools will be misapplied, for example, by incorrectly labeling legitimate content as false or by being employed in politically or socially biased ways. Overreliance on automated systems is another growing concern, as they may miss edge cases or fail silently in unfamiliar situations.

\section*{Reproducibility Statement} 
We are committed to ensuring the reproducibility of our results. To this end, we will release the full source code and evaluation scripts upon publication. Our paper clearly and fully describes the proposed feature extraction method and the model architecture in Section \ref{sec:method}, and provides comprehensive details on the experimental setup in Section \ref{sec:experimental_setup}, including used datasets, preprocessing steps, training configurations, and hyperparameters. Additionally, ablations and variant evaluations in Section \ref{sec:ablation} and Appendix \ref{appendix:ablations} further support reproducibility. 


\bibliography{iclr2026_conference}

\begin{thebibliography}{70}
\providecommand{\natexlab}[1]{#1}
\providecommand{\url}[1]{\texttt{#1}}
\expandafter\ifx\csname urlstyle\endcsname\relax
  \providecommand{\doi}[1]{doi: #1}\else
  \providecommand{\doi}{doi: \begingroup \urlstyle{rm}\Url}\fi

\bibitem[Bai et~al.(2024)Bai, Liu, Zhang, Zhang, Wang, Peng, Hu, and Li]{noise2}
Weiming Bai, Yufan Liu, Zhipeng Zhang, Xinyi Zhang, Bo~Wang, Chengwei Peng, Weiming Hu, and Bing Li.
\newblock Learn from noise: Detecting deepfakes via regional noise consistency.
\newblock In \emph{2024 International Joint Conference on Neural Networks (IJCNN)}, pp.\  1--8. IEEE, 2024.

\bibitem[{Black Forest Labs}(2025)]{flux}
{Black Forest Labs}.
\newblock Flux. 1 kontext: Flow matching for in-context image generation and editing in latent space.
\newblock \emph{arXiv preprint arXiv:2506.15742}, 2025.

\bibitem[Brock et~al.(2018)Brock, Donahue, and Simonyan]{biggan}
Andrew Brock, Jeff Donahue, and Karen Simonyan.
\newblock Large scale gan training for high fidelity natural image synthesis.
\newblock \emph{arXiv preprint arXiv:1809.11096}, 2018.

\bibitem[Chen et~al.(2024)Chen, Zeng, Yang, and Yang]{drct}
Baoying Chen, Jishen Zeng, Jianquan Yang, and Rui Yang.
\newblock Drct: Diffusion reconstruction contrastive training towards universal detection of diffusion generated images.
\newblock In \emph{Forty-first International Conference on Machine Learning}, 2024.

\bibitem[Cheng et~al.(2025)Cheng, Lyu, Wang, Zhang, and Sehwag]{cospy}
Siyuan Cheng, Lingjuan Lyu, Zhenting Wang, Xiangyu Zhang, and Vikash Sehwag.
\newblock {CO-SPY}: Combining semantic and pixel features to detect synthetic images by ai.
\newblock In \emph{Proceedings of the Computer Vision and Pattern Recognition Conference}, pp.\  13455--13465, 2025.

\bibitem[Choi et~al.(2018)Choi, Choi, Kim, Ha, Kim, and Choo]{stargan}
Yunjey Choi, Minje Choi, Munyoung Kim, Jung-Woo Ha, Sunghun Kim, and Jaegul Choo.
\newblock Stargan: Unified generative adversarial networks for multi-domain image-to-image translation.
\newblock In \emph{Proceedings of the IEEE conference on computer vision and pattern recognition}, pp.\  8789--8797, 2018.

\bibitem[Chu et~al.(2024)Chu, Xu, Wang, Zhang, You, and Zhou]{fire}
Beilin Chu, Xuan Xu, Xin Wang, Yufei Zhang, Weike You, and Linna Zhou.
\newblock Fire: Robust detection of diffusion-generated images via frequency-guided reconstruction error.
\newblock \emph{arXiv preprint arXiv:2412.07140}, 2024.

\bibitem[Cozzolino et~al.(2024)Cozzolino, Poggi, Corvi, Nie{\ss}ner, and Verdoliva]{raisingthebar}
Davide Cozzolino, Giovanni Poggi, Riccardo Corvi, Matthias Nie{\ss}ner, and Luisa Verdoliva.
\newblock Raising the bar of ai-generated image detection with clip.
\newblock In \emph{Proceedings of the IEEE/CVF Conference on Computer Vision and Pattern Recognition}, pp.\  4356--4366, 2024.

\bibitem[de~Rancourt-Raymond \& Smaili(2023)de~Rancourt-Raymond and Smaili]{de2023unethical}
Audrey de~Rancourt-Raymond and Nadia Smaili.
\newblock The unethical use of deepfakes.
\newblock \emph{Journal of Financial Crime}, 30\penalty0 (4):\penalty0 1066--1077, 2023.

\bibitem[Delfino(2022)]{delfino2022deepfakes}
Rebecca~A Delfino.
\newblock Deepfakes on trial: a call to expand the trial judge's gatekeeping role to protect legal proceedings from technological fakery.
\newblock \emph{Hastings LJ}, 74:\penalty0 293, 2022.

\bibitem[Dhariwal \& Nichol(2021)Dhariwal and Nichol]{adm}
Prafulla Dhariwal and Alexander Nichol.
\newblock Diffusion models beat {GANs} on image synthesis.
\newblock \emph{Advances in neural information processing systems}, 34:\penalty0 8780--8794, 2021.

\bibitem[Esser et~al.(2021)Esser, Rombach, and Ommer]{taming}
Patrick Esser, Robin Rombach, and Bjorn Ommer.
\newblock Taming transformers for high-resolution image synthesis.
\newblock In \emph{Proceedings of the IEEE/CVF conference on computer vision and pattern recognition}, pp.\  12873--12883, 2021.

\bibitem[Frank et~al.(2020)Frank, Eisenhofer, Sch{\"o}nherr, Fischer, Kolossa, and Holz]{frank2020leveraging}
Joel Frank, Thorsten Eisenhofer, Lea Sch{\"o}nherr, Asja Fischer, Dorothea Kolossa, and Thorsten Holz.
\newblock Leveraging frequency analysis for deep fake image recognition.
\newblock In \emph{International conference on machine learning}, pp.\  3247--3258. PMLR, 2020.

\bibitem[Goodfellow et~al.(2020)Goodfellow, Pouget-Abadie, Mirza, Xu, Warde-Farley, Ozair, Courville, and Bengio]{gan}
Ian Goodfellow, Jean Pouget-Abadie, Mehdi Mirza, Bing Xu, David Warde-Farley, Sherjil Ozair, Aaron Courville, and Yoshua Bengio.
\newblock Generative adversarial networks.
\newblock \emph{Communications of the ACM}, 63\penalty0 (11):\penalty0 139--144, 2020.

\bibitem[Gu et~al.(2022)Gu, Chen, Bao, Wen, Zhang, Chen, Yuan, and Guo]{vqdm}
Shuyang Gu, Dong Chen, Jianmin Bao, Fang Wen, Bo~Zhang, Dongdong Chen, Lu~Yuan, and Baining Guo.
\newblock Vector quantized diffusion model for text-to-image synthesis.
\newblock In \emph{Proceedings of the IEEE/CVF conference on computer vision and pattern recognition}, pp.\  10696--10706, 2022.

\bibitem[He et~al.(2016)He, Zhang, Ren, and Sun]{rn50}
Kaiming He, Xiangyu Zhang, Shaoqing Ren, and Jian Sun.
\newblock Deep residual learning for image recognition.
\newblock In \emph{Proceedings of the IEEE conference on computer vision and pattern recognition}, pp.\  770--778, 2016.

\bibitem[Ho \& Salimans(2022)Ho and Salimans]{cfguidance}
Jonathan Ho and Tim Salimans.
\newblock Classifier-free diffusion guidance.
\newblock \emph{arXiv preprint arXiv:2207.12598}, 2022.

\bibitem[Ho et~al.(2020)Ho, Jain, and Abbeel]{ddpm}
Jonathan Ho, Ajay Jain, and Pieter Abbeel.
\newblock Denoising diffusion probabilistic models.
\newblock \emph{Advances in neural information processing systems}, 33:\penalty0 6840--6851, 2020.

\bibitem[Ilharco et~al.(2021)Ilharco, Wortsman, Wightman, Gordon, Carlini, Taori, Dave, Shankar, Namkoong, Miller, Hajishirzi, Farhadi, and Schmidt]{ilharco_gabriel_2021_5143773}
Gabriel Ilharco, Mitchell Wortsman, Ross Wightman, Cade Gordon, Nicholas Carlini, Rohan Taori, Achal Dave, Vaishaal Shankar, Hongseok Namkoong, John Miller, Hannaneh Hajishirzi, Ali Farhadi, and Ludwig Schmidt.
\newblock Openclip, July 2021.
\newblock URL \url{https://doi.org/10.5281/zenodo.5143773}.
\newblock If you use this software, please cite it as below.

\bibitem[Karras et~al.(2017)Karras, Aila, Laine, and Lehtinen]{karras2017progressive}
Tero Karras, Timo Aila, Samuli Laine, and Jaakko Lehtinen.
\newblock Progressive growing of gans for improved quality, stability, and variation.
\newblock \emph{arXiv preprint arXiv:1710.10196}, 2017.

\bibitem[Karras et~al.(2022)Karras, Aittala, Aila, and Laine]{dpm_solver}
Tero Karras, Miika Aittala, Timo Aila, and Samuli Laine.
\newblock Elucidating the design space of diffusion-based generative models.
\newblock \emph{Advances in neural information processing systems}, 35:\penalty0 26565--26577, 2022.

\bibitem[Kingma et~al.(2013)Kingma, Welling, et~al.]{vae}
Diederik~P Kingma, Max Welling, et~al.
\newblock Auto-encoding variational bayes, 2013.

\bibitem[Koutras \& Selvadurai(2024)Koutras and Selvadurai]{KoutrasSelvadurai2024}
N.~Koutras and N.~Selvadurai (eds.).
\newblock \emph{Recreating Creativity, Reinventing Inventiveness: AI and Intellectual Property Law}.
\newblock Routledge, 1st edition, 2024.
\newblock \doi{10.4324/9781003260127}.
\newblock URL \url{https://doi.org/10.4324/9781003260127}.

\bibitem[Li et~al.(2024)Li, Zhu, Fu, Guo, Liu, Yang, Liu, and Zha]{noiseassisted}
Dong Li, Jiaying Zhu, Xueyang Fu, Xun Guo, Yidi Liu, Gang Yang, Jiawei Liu, and Zheng-Jun Zha.
\newblock Noise-assisted prompt learning for image forgery detection and localization.
\newblock In \emph{European Conference on Computer Vision}, pp.\  18--36. Springer, 2024.

\bibitem[Liu et~al.(2021)Liu, Lin, Cao, Hu, Wei, Zhang, Lin, and Guo]{swint}
Ze~Liu, Yutong Lin, Yue Cao, Han Hu, Yixuan Wei, Zheng Zhang, Stephen Lin, and Baining Guo.
\newblock Swin transformer: Hierarchical vision transformer using shifted windows.
\newblock In \emph{Proceedings of the IEEE/CVF international conference on computer vision}, pp.\  10012--10022, 2021.

\bibitem[Liu et~al.(2020)Liu, Qi, and Torr]{gramnet}
Zhengzhe Liu, Xiaojuan Qi, and Philip~HS Torr.
\newblock Global texture enhancement for fake face detection in the wild.
\newblock In \emph{Proceedings of the IEEE/CVF conference on computer vision and pattern recognition}, pp.\  8060--8069, 2020.

\bibitem[Liu et~al.(2022)Liu, Mao, Wu, Feichtenhofer, Darrell, and Xie]{convnext}
Zhuang Liu, Hanzi Mao, Chao-Yuan Wu, Christoph Feichtenhofer, Trevor Darrell, and Saining Xie.
\newblock A convnet for the 2020s.
\newblock In \emph{Proceedings of the IEEE/CVF conference on computer vision and pattern recognition}, pp.\  11976--11986, 2022.

\bibitem[Liy \& InIctuOculi(2018)Liy and InIctuOculi]{eyeblink}
Chang~M Liy and LYUS InIctuOculi.
\newblock Exposing ai created fake videos by detecting eye blinking.
\newblock In \emph{2018 IEEE InterG national Workshop on Information Forensics and Security (WIFS). IEEE}, 2018.

\bibitem[Loshchilov \& Hutter(2017)Loshchilov and Hutter]{loshchilov2017decoupled}
Ilya Loshchilov and Frank Hutter.
\newblock Decoupled weight decay regularization.
\newblock \emph{International Conference on Learning Representations}, 2017.

\bibitem[Luo et~al.(2023)Luo, Tan, Huang, Li, and Zhao]{lcm}
Simian Luo, Yiqin Tan, Longbo Huang, Jian Li, and Hang Zhao.
\newblock Latent consistency models: Synthesizing high-resolution images with few-step inference.
\newblock \emph{arXiv preprint arXiv:2310.04378}, 2023.

\bibitem[Luo et~al.(2021)Luo, Zhang, Yan, and Liu]{luo2021generalizing}
Yuchen Luo, Yong Zhang, Junchi Yan, and Wei Liu.
\newblock Generalizing face forgery detection with high-frequency features.
\newblock In \emph{Proceedings of the IEEE/CVF conference on computer vision and pattern recognition}, pp.\  16317--16326, 2021.

\bibitem[Luo et~al.(2024)Luo, Du, Yan, and Ding]{lare}
Yunpeng Luo, Junlong Du, Ke~Yan, and Shouhong Ding.
\newblock Lare\^{} 2: Latent reconstruction error based method for diffusion-generated image detection.
\newblock In \emph{Proceedings of the IEEE/CVF Conference on Computer Vision and Pattern Recognition}, pp.\  17006--17015, 2024.

\bibitem[Ma et~al.(2023)Ma, Duan, Kong, Shi, and Xu]{sedid}
Ruipeng Ma, Jinhao Duan, Fei Kong, Xiaoshuang Shi, and Kaidi Xu.
\newblock Exposing the fake: Effective diffusion-generated images detection.
\newblock \emph{arXiv preprint arXiv:2307.06272}, 2023.

\bibitem[{Midjourney}(2024)]{midjourney}
{Midjourney}.
\newblock {Midjourney}.
\newblock 2024.
\newblock URL \url{https://www.midjourney.com/home}.

\bibitem[Nichol et~al.(2021)Nichol, Dhariwal, Ramesh, Shyam, Mishkin, McGrew, Sutskever, and Chen]{glide}
Alex Nichol, Prafulla Dhariwal, Aditya Ramesh, Pranav Shyam, Pamela Mishkin, Bob McGrew, Ilya Sutskever, and Mark Chen.
\newblock Glide: Towards photorealistic image generation and editing with text-guided diffusion models.
\newblock \emph{arXiv preprint arXiv:2112.10741}, 2021.

\bibitem[Ojha et~al.(2023)Ojha, Li, and Lee]{univfd}
Utkarsh Ojha, Yuheng Li, and Yong~Jae Lee.
\newblock Towards universal fake image detectors that generalize across generative models.
\newblock In \emph{Proceedings of the IEEE/CVF Conference on Computer Vision and Pattern Recognition}, pp.\  24480--24489, 2023.

\bibitem[Park et~al.(2019)Park, Liu, Wang, and Zhu]{spade}
Taesung Park, Ming-Yu Liu, Ting-Chun Wang, and Jun-Yan Zhu.
\newblock Semantic image synthesis with spatially-adaptive normalization.
\newblock In \emph{Proceedings of the IEEE/CVF conference on computer vision and pattern recognition}, pp.\  2337--2346, 2019.

\bibitem[Parmar et~al.(2018)Parmar, Vaswani, Uszkoreit, Kaiser, Shazeer, Ku, and Tran]{imagetransformer}
Niki Parmar, Ashish Vaswani, Jakob Uszkoreit, Lukasz Kaiser, Noam Shazeer, Alexander Ku, and Dustin Tran.
\newblock Image transformer.
\newblock In \emph{International conference on machine learning}, pp.\  4055--4064. PMLR, 2018.

\bibitem[Podell et~al.(2023)Podell, English, Lacey, Blattmann, Dockhorn, M{\"u}ller, Penna, and Rombach]{sdxl}
Dustin Podell, Zion English, Kyle Lacey, Andreas Blattmann, Tim Dockhorn, Jonas M{\"u}ller, Joe Penna, and Robin Rombach.
\newblock Sdxl: Improving latent diffusion models for high-resolution image synthesis.
\newblock \emph{arXiv preprint arXiv:2307.01952}, 2023.

\bibitem[Qian et~al.(2020)Qian, Yin, Sheng, Chen, and Shao]{f3net}
Yuyang Qian, Guojun Yin, Lu~Sheng, Zixuan Chen, and Jing Shao.
\newblock Thinking in frequency: Face forgery detection by mining frequency-aware clues.
\newblock In \emph{European conference on computer vision}, pp.\  86--103. Springer, 2020.

\bibitem[Radford et~al.(2021)Radford, Kim, Hallacy, Ramesh, Goh, Agarwal, Sastry, Askell, Mishkin, Clark, et~al.]{radford2021learning}
Alec Radford, Jong~Wook Kim, Chris Hallacy, Aditya Ramesh, Gabriel Goh, Sandhini Agarwal, Girish Sastry, Amanda Askell, Pamela Mishkin, Jack Clark, et~al.
\newblock Learning transferable visual models from natural language supervision.
\newblock In \emph{International conference on machine learning}, pp.\  8748--8763. PmLR, 2021.

\bibitem[Ramesh et~al.(2021)Ramesh, Pavlov, Goh, Gray, Voss, Radford, Chen, and Sutskever]{zeroshot}
Aditya Ramesh, Mikhail Pavlov, Gabriel Goh, Scott Gray, Chelsea Voss, Alec Radford, Mark Chen, and Ilya Sutskever.
\newblock Zero-shot text-to-image generation.
\newblock In \emph{International conference on machine learning}, pp.\  8821--8831. Pmlr, 2021.

\bibitem[Ricker et~al.(2024)Ricker, Lukovnikov, and Fischer]{ricker2024aeroblade}
Jonas Ricker, Denis Lukovnikov, and Asja Fischer.
\newblock Aeroblade: Training-free detection of latent diffusion images using autoencoder reconstruction error.
\newblock In \emph{Proceedings of the IEEE/CVF Conference on Computer Vision and Pattern Recognition}, pp.\  9130--9140, 2024.

\bibitem[Rombach et~al.(2022{\natexlab{a}})Rombach, Blattmann, Lorenz, Esser, and Ommer]{ldm}
Robin Rombach, Andreas Blattmann, Dominik Lorenz, Patrick Esser, and Bj{\"o}rn Ommer.
\newblock High-resolution image synthesis with latent diffusion models.
\newblock In \emph{Proceedings of the IEEE/CVF conference on computer vision and pattern recognition}, pp.\  10684--10695, 2022{\natexlab{a}}.

\bibitem[Rombach et~al.(2022{\natexlab{b}})Rombach, Blattmann, Lorenz, Esser, and Ommer]{sdv}
Robin Rombach, Andreas Blattmann, Dominik Lorenz, Patrick Esser, and Bj{\"o}rn Ommer.
\newblock High-resolution image synthesis with latent diffusion models.
\newblock In \emph{Proceedings of the IEEE/CVF conference on computer vision and pattern recognition}, pp.\  10684--10695, 2022{\natexlab{b}}.

\bibitem[Rossler et~al.(2019)Rossler, Cozzolino, Verdoliva, Riess, Thies, and Nie{\ss}ner]{faceforensics}
Andreas Rossler, Davide Cozzolino, Luisa Verdoliva, Christian Riess, Justus Thies, and Matthias Nie{\ss}ner.
\newblock Faceforensics++: Learning to detect manipulated facial images.
\newblock In \emph{Proceedings of the IEEE/CVF international conference on computer vision}, pp.\  1--11, 2019.

\bibitem[Saharia et~al.(2022)Saharia, Chan, Saxena, Li, Whang, Denton, Ghasemipour, Gontijo~Lopes, Karagol~Ayan, Salimans, et~al.]{imagen}
Chitwan Saharia, William Chan, Saurabh Saxena, Lala Li, Jay Whang, Emily~L Denton, Kamyar Ghasemipour, Raphael Gontijo~Lopes, Burcu Karagol~Ayan, Tim Salimans, et~al.
\newblock Photorealistic text-to-image diffusion models with deep language understanding.
\newblock \emph{Advances in neural information processing systems}, 35:\penalty0 36479--36494, 2022.

\bibitem[Sandoval et~al.(2024)Sandoval, de~Almeida~Vau, Solaas, and Rodrigues]{sandoval2024threat}
Maria-Paz Sandoval, Maria de~Almeida~Vau, John Solaas, and Luano Rodrigues.
\newblock Threat of deepfakes to the criminal justice system: a systematic review.
\newblock \emph{Crime Science}, 13\penalty0 (1):\penalty0 41, 2024.

\bibitem[Sohn et~al.(2015)Sohn, Lee, and Yan]{sohn2015learning}
Kihyuk Sohn, Honglak Lee, and Xinchen Yan.
\newblock Learning structured output representation using deep conditional generative models.
\newblock \emph{Advances in neural information processing systems}, 28, 2015.

\bibitem[Song et~al.(2020)Song, Meng, and Ermon]{ddim}
Jiaming Song, Chenlin Meng, and Stefano Ermon.
\newblock Denoising diffusion implicit models.
\newblock \emph{arXiv preprint arXiv:2010.02502}, 2020.

\bibitem[Touvron et~al.(2021)Touvron, Cord, Douze, Massa, Sablayrolles, and J{\'e}gou]{deits}
Hugo Touvron, Matthieu Cord, Matthijs Douze, Francisco Massa, Alexandre Sablayrolles, and Herv{\'e} J{\'e}gou.
\newblock Training data-efficient image transformers \& distillation through attention.
\newblock In \emph{International conference on machine learning}, pp.\  10347--10357. PMLR, 2021.

\bibitem[Twomey et~al.(2023)Twomey, Ching, Aylett, Quayle, Linehan, and Murphy]{twomey2023deepfake}
John Twomey, Didier Ching, Matthew~Peter Aylett, Michael Quayle, Conor Linehan, and Gillian Murphy.
\newblock Do deepfake videos undermine our epistemic trust? a thematic analysis of tweets that discuss deepfakes in the russian invasion of ukraine.
\newblock \emph{Plos one}, 18\penalty0 (10):\penalty0 e0291668, 2023.

\bibitem[Van~den Oord et~al.(2016)Van~den Oord, Kalchbrenner, Espeholt, Vinyals, Graves, et~al.]{pixelcnn}
Aaron Van~den Oord, Nal Kalchbrenner, Lasse Espeholt, Oriol Vinyals, Alex Graves, et~al.
\newblock Conditional image generation with pixelcnn decoders.
\newblock \emph{Advances in neural information processing systems}, 29, 2016.

\bibitem[Van Den~Oord et~al.(2017)Van Den~Oord, Vinyals, et~al.]{vandenoord}
Aaron Van Den~Oord, Oriol Vinyals, et~al.
\newblock Neural discrete representation learning.
\newblock \emph{Advances in neural information processing systems}, 30, 2017.

\bibitem[van~der Maaten \& Hinton(2008)van~der Maaten and Hinton]{vandermaaten08a}
Laurens van~der Maaten and Geoffrey Hinton.
\newblock Visualizing data using t-sne.
\newblock \emph{Journal of Machine Learning Research}, 9\penalty0 (86):\penalty0 2579--2605, 2008.
\newblock URL \url{http://jmlr.org/papers/v9/vandermaaten08a.html}.

\bibitem[Wang et~al.(2020)Wang, Wang, Zhang, Owens, and Efros]{wang2020cnn}
Sheng-Yu Wang, Oliver Wang, Richard Zhang, Andrew Owens, and Alexei~A Efros.
\newblock Cnn-generated images are surprisingly easy to spot... for now.
\newblock In \emph{Proceedings of the IEEE/CVF conference on computer vision and pattern recognition}, pp.\  8695--8704, 2020.

\bibitem[Wang \& Chow(2023)Wang and Chow]{noise1}
Tianyi Wang and Kam~Pui Chow.
\newblock Noise based deepfake detection via multi-head relative-interaction.
\newblock In \emph{Proceedings of the AAAI Conference on Artificial Intelligence}, volume~37, pp.\  14548--14556, 2023.

\bibitem[Wang et~al.(2023)Wang, Bao, Zhou, Wang, Hu, Chen, and Li]{dire}
Zhendong Wang, Jianmin Bao, Wengang Zhou, Weilun Wang, Hezhen Hu, Hong Chen, and Houqiang Li.
\newblock Dire for diffusion-generated image detection.
\newblock In \emph{Proceedings of the IEEE/CVF International Conference on Computer Vision}, pp.\  22445--22455, 2023.

\bibitem[{Wukong}(2024)]{wukong}
{Wukong}.
\newblock Wukong.
\newblock 2024.
\newblock URL \url{https://xihe.mindspore.cn/modelzoo}.

\bibitem[Yan et~al.(2025)Yan, Li, Cai, Hao, Jiang, Hu, and Xie]{aide}
Shilin Yan, Ouxiang Li, Jiayin Cai, Yanbin Hao, Xiaolong Jiang, Yao Hu, and Weidi Xie.
\newblock A sanity check for {AI}-generated image detection.
\newblock \emph{ICLR}, 2025.

\bibitem[Yang et~al.(2019)Yang, Li, and Lyu]{headpose}
Xin Yang, Yuezun Li, and Siwei Lyu.
\newblock Exposing deep fakes using inconsistent head poses.
\newblock In \emph{ICASSP 2019-2019 IEEE international conference on acoustics, speech and signal processing (ICASSP)}, pp.\  8261--8265. IEEE, 2019.

\bibitem[Zhang et~al.(2023)Zhang, Rao, and Agrawala]{controlnet}
Lvmin Zhang, Anyi Rao, and Maneesh Agrawala.
\newblock Adding conditional control to text-to-image diffusion models.
\newblock In \emph{Proceedings of the IEEE/CVF international conference on computer vision}, pp.\  3836--3847, 2023.

\bibitem[Zhang et~al.(2019)Zhang, Karaman, and Chang]{zhang2019detecting}
Xu~Zhang, Svebor Karaman, and Shih-Fu Chang.
\newblock Detecting and simulating artifacts in gan fake images.
\newblock In \emph{2019 IEEE international workshop on information forensics and security (WIFS)}, pp.\  1--6. IEEE, 2019.

\bibitem[Zhang et~al.(2024)Zhang, Wang, Yu, Gao, Shen, and Chen]{mfclip}
Yaning Zhang, Tianyi Wang, Zitong Yu, Zan Gao, Linlin Shen, and Shengyong Chen.
\newblock Mfclip: Multi-modal fine-grained clip for generalizable diffusion face forgery detection.
\newblock \emph{arXiv preprint arXiv:2409.09724}, 2024.

\bibitem[Zhang \& Xu(2023)Zhang and Xu]{dnf}
Yichi Zhang and Xiaogang Xu.
\newblock Diffusion noise feature: Accurate and fast generated image detection.
\newblock \emph{arXiv preprint arXiv:2312.02625}, 2023.

\bibitem[Zhao et~al.(2017)Zhao, Song, and Ermon]{infovae}
Shengjia Zhao, Jiaming Song, and Stefano Ermon.
\newblock Infovae: Information maximizing variational autoencoders.
\newblock \emph{arXiv preprint arXiv:1706.02262}, 2017.

\bibitem[Zhong et~al.(2023)Zhong, Xu, Li, Qian, and Zhang]{patchcraft}
Nan Zhong, Yiran Xu, Sheng Li, Zhenxing Qian, and Xinpeng Zhang.
\newblock Patchcraft: Exploring texture patch for efficient ai-generated image detection.
\newblock \emph{arXiv preprint arXiv:2311.12397}, 2023.

\bibitem[Zhu et~al.(2017)Zhu, Park, Isola, and Efros]{zhu2017unpaired}
Jun-Yan Zhu, Taesung Park, Phillip Isola, and Alexei~A Efros.
\newblock Unpaired image-to-image translation using cycle-consistent adversarial networks.
\newblock In \emph{Proceedings of the IEEE international conference on computer vision}, pp.\  2223--2232, 2017.

\bibitem[Zhu et~al.(2023{\natexlab{a}})Zhu, Chen, Huang, Li, Hu, Hu, and Wang]{zhu2023gendet}
Mingjian Zhu, Hanting Chen, Mouxiao Huang, Wei Li, Hailin Hu, Jie Hu, and Yunhe Wang.
\newblock Gen{D}et: Towards good generalizations for ai-generated image detection.
\newblock \emph{arXiv preprint arXiv:2312.08880}, 2023{\natexlab{a}}.

\bibitem[Zhu et~al.(2023{\natexlab{b}})Zhu, Chen, Yan, Huang, Lin, Li, Tu, Hu, Hu, and Wang]{genimage}
Mingjian Zhu, Hanting Chen, Qiangyu Yan, Xudong Huang, Guanyu Lin, Wei Li, Zhijun Tu, Hailin Hu, Jie Hu, and Yunhe Wang.
\newblock Genimage: A million-scale benchmark for detecting ai-generated image.
\newblock \emph{Advances in Neural Information Processing Systems}, 36:\penalty0 77771--77782, 2023{\natexlab{b}}.

\end{thebibliography}
\bibliographystyle{iclr2026_conference}

\newpage

\appendix
\section*{Appendix}
The appendix consists of the following sections: \ref{appendix:ablations}. Additional Ablation Studies, \ref{appendix:traj_analysis}. Latent Trajectory Spatial Analysis, \ref{appendix:complete_acc}. Complete Accuracy and AP on GenImage, \ref{appendix:cls_pooling}. Architectural Details of the CLS-pooling, and \ref{appendix:embddings}. Embedding Space Analysis.
\section{Additional ablation studies}
\label{appendix:ablations}
To further understand the key design choices and components of the LATTE framework, we conduct a series of additional ablation studies. 
All ablation results reported in this section are based on models trained using the SDv1.4 subset of GenImage \citep{genimage}.

\subsection{Benefit of average pooling} Standard pooling in LATTE assumes equal importance across all timesteps in the trajectory. To test this design choice, we experiment with a weighted pooling mechanism that assigns importance scores to each timestep using a linear gating function and softmax normalization. As shown in Table \ref{tab:weighting}, this approach performs worse than simple average pooling - suggesting that all steps provide equally informative signals.
We also consider CLS pooling, where a special token aggregates the sequence of latents through self-attention with positional encodings.
The goal is to assess whether allowing the latents to refine each other via self-attention and incorporating sequence order can improve performance. This variant slightly underperforms, suggesting that LATTE is already expressive enough without additional attention-based aggregations.

\vspace{-0.2cm}
\begin{table}[h]
\centering
\caption{\textbf{Accuracy(\%) comparison between different aggregation strategies.} Average pooling outperforms learnable weighted pooling by 9.4\% and CLS pooling by 1.4\%.}
\label{tab:weighting}
\resizebox{\textwidth}{!}{
\begin{tabular}{l|cccccccc|c}
\toprule
\textbf{Configuration} & Midjourney & SDv1.4 & SDv1.5 & ADM & GLIDE & Wukong & VQDM & BigGAN & \textit{\textbf{Avg.}} \\
\toprule
Weighted pooling & 73.0 & 99.8 & 99.7 & 72.0 & 79.0 & 98.7 & 74.4 & 56.3 & 81.6 \\
CLS pooling & 85.8 & 99.0 & 99.0 & \textbf{83.2} & 86.7 & 96.8 & \textbf{88.7} & 77.8 & 89.6 \\
\rowcolor{yellow!15}
\textbf{Average pooling} & \textbf{88.8} & \textbf{100} & \textbf{99.9} & 74.0 & \textbf{95.8} & \textbf{98.9} & 80.8 & \textbf{89.8} & \textbf{91.0} \\
\bottomrule
\end{tabular}
}
\end{table}

\subsection{Effect of latent extraction configuration}
The sequence of latents is obtained by first encoding real and fake images into latent space using a frozen VAE, followed by partial reconstruction via a pre-trained diffusion model. At each timestep, noise is added to the VAE latents and then partially denoised via the UNet, capturing intermediate latent representations along the reconstruction trajectory.

We ablate two factors in this latent extraction pipeline: the choice of sampling method (DDPM vs. DDIM, Table \ref{tab:noise_scheduler}) and the choice of U-Net backbone (Stable Diffusion v1.5 vs. v2.1, Table \ref{tab:unet_backbone}). 
For the sampling method, we use Stable Diffusion v2.1 as the backbone, while for the U-Net model comparison, we fix the scheduler to DDPM.

\begin{table}[h]
\centering
\caption{\textbf{Accuracy(\%) comparison between DDPM and DDIM-based latent extraction.} DDPM improved accuracy by 7.2\%.}
\label{tab:noise_scheduler}
\resizebox{\textwidth}{!}{
\begin{tabular}{l|cccccccc|c}
\toprule
\textbf{Sampling method} & Midjourney & SDv1.4 & SDv1.5 & ADM & GLIDE & Wukong & VQDM & BigGAN & \textit{\textbf{Avg.}} \\
\toprule
DDIM & 77.0 & 99.7 & 99.5 & \textbf{74.6} & 82.2 & 98.2 & 77.4 & 61.8 & 83.8 \\
\rowcolor{yellow!15}
\textbf{DDPM} & \textbf{88.8} & \textbf{100} & \textbf{99.9} & 74.0 & \textbf{95.8} & \textbf{98.9} & \textbf{80.8} & \textbf{89.8} & \textbf{91.0} \\
\bottomrule
\end{tabular}
}
\end{table}

\begin{table}[h]
\centering
\caption{\textbf{Accuracy(\%) comparison between SDv1.5 and SDv2.1-based latent extraction.} SDv2.1 improves accuracy by 3.9\%.}
\label{tab:unet_backbone}
\resizebox{\textwidth}{!}{
\begin{tabular}{l|cccccccc|c}
\toprule
\textbf{U-Net Backbone} & Midjourney & SDv1.4 & SDv1.5 & ADM & GLIDE & Wukong & VQDM & BigGAN & \textit{\textbf{Avg.}} \\
\toprule
SDv1.5 & 81.7 & 99.7 & 99.5 & \textbf{77.7} & 90.6 & 98.0 & 78.1 & 71.5 & 87.1 \\
\rowcolor{yellow!15}
\textbf{SDv2.1} & \textbf{88.8} & \textbf{100} & \textbf{99.9} & 74.0 & \textbf{95.8} & \textbf{98.9} & \textbf{80.8} & \textbf{89.8} & \textbf{91.0} \\
\bottomrule
\end{tabular}
}
\end{table}

The results indicate that LATTE’s performance is sensitive to the latent extraction configuration, highlighting the importance of both the sampling method and the U-Net backbone. Switching from DDIM to DDPM yields a substantial improvement in average detection accuracy (+7.2\%), with particularly large gains on datasets such as Midjourney and BigGAN. This suggests that the stochastic denoising dynamics captured by DDPM produce richer latent trajectories, enhancing the discriminative signal between real and generated images. Similarly, upgrading the U-Net backbone from SDv1.5 to SDv2.1 further improves average accuracy (+3.9\%), reflecting the impact of more expressive latent representations on the model’s ability to capture subtle generative artifacts. While some datasets, such as ADM, show minimal changes, likely due to inherent detection difficulty or saturation effects, the overall trend confirms that both the scheduler and backbone play complementary roles: the scheduler shapes the temporal evolution of latents, whereas the backbone determines the quality of the underlying feature space. Despite these variations, LATTE maintains high and consistent performance across all configurations, demonstrating its robustness and reliability as a diffusion-generated image detector.

\subsection{Influence of vision backbone fine-tuning}
Our default setup fine-tunes the vision encoder. To quantify the added benefit of this choice, we compare against a variant where we freeze the backbone and train only the LATTE-specific components. Table \ref{tab:finetune} reports per-generator accuracy for both settings. We observe major improvements for both vision backbones when fine-tuned, with 15.5\% accuracy gain for CLIP ViT-L/14 \citep{radford2021learning} and 9\% for ConvNeXt \citep{convnext}. This likely stems from the fact that frozen backbones retain features that were never explicitly optimized for real vs. fake discrimination, leading to an embedding space that is misaligned with the objectives of generated image detection. Without adaptation, our model may struggle to effectively ground latent trajectories in meaningful visual semantics. Fine-tuning, by contrast, enables the backbone to specialize its representations for this task, enhancing the alignment between visual and latent features essential for robust detection.

\vspace{-0.3em}
\begin{table}[h]
  \centering
  \caption{\textbf{Accuracy (\%) comparison for different vision backbones and fine-tuning vs. frozen settings on the GenImage dataset.} Fine-tuned ConvNeXt yields the best performance.}
  \label{tab:finetune}
  \resizebox{\columnwidth}{!}{%
    \begin{tabular}{llcccccccc|c}
      \toprule
      \textbf{Backbone} & \textbf{Setting} 
        & Midjourney & SDv1.4 & SDv1.5 & ADM & GLIDE & Wukong & VQDM & BigGAN & \textit{\textbf{Avg.}} \\
      \midrule
      \multirow{2}{*}{CLIP ViT-L/14}
      & Frozen  & 60.3 & 99.8 & 99.9 & 53.6 & 50.3 & 99.8 & 51.4 & 50.1 & 70.6  \\
      & Fine-tuned
        & \cellcolor{yellow!15}\textbf{98.2}
        & \cellcolor{yellow!15}99.9
        & \cellcolor{yellow!15}\textbf{100} 
        & \cellcolor{yellow!15}68.4
        & \cellcolor{yellow!15}91.8
        & \cellcolor{yellow!15}\textbf{99.9}
        & \cellcolor{yellow!15}58.9
        & \cellcolor{yellow!15}72.1
        & \cellcolor{yellow!15}86.1 \\
        \midrule
      \multirow{2}{*}{ConvNeXt}
      & Frozen  & 79.7 & 99.3 & 99.1 & 64.4 & 74.2 & 95.9 & 72.4 & 70.9 & 81.9  \\
      & Fine-tuned
        & \cellcolor{yellow!15}88.8
        & \cellcolor{yellow!15}\textbf{100}
        & \cellcolor{yellow!15}99.9 
        & \cellcolor{yellow!15}\textbf{74.0}
        & \cellcolor{yellow!15}\textbf{95.8}
        & \cellcolor{yellow!15}98.9
        & \cellcolor{yellow!15}\textbf{80.8}
        & \cellcolor{yellow!15}\textbf{89.8}
        & \cellcolor{yellow!15}\textbf{91.0}\\
    \bottomrule
    \end{tabular}%
  }
\end{table}

\subsection{Influence of separate latent processing strategy}
The default LATTE architecture, as described in Section 3, processes the latent trajectory by refining each timestep independently using a dedicated transformer decoder. An alternative approach is to stack the latent embeddings from all timesteps into a single sequence and process them jointly through a shared transformer decoder stack, enforcing full parameter sharing across the sequence. As shown in Table \ref{tab:latent_strategy}, decoding each timestep separately achieves higher accuracy across most generators, suggesting that preserving per-timestep decoding helps the model retain specific features from the denoising trajectory. 

\begin{table}[h]
\centering
\caption{\textbf{Accuracy(\%) comparison} between separate vs. joint latent processing strategies. Processing timesteps separately yields the highest average accuracy, outperforming joint processing by 0.8\%.}
\label{tab:latent_strategy}
\resizebox{\textwidth}{!}{
\begin{tabular}{lccccccccc}
\toprule
\textbf{Latent strategy} & Midjourney & SDv1.4 & SDv1.5 & ADM & GLIDE & Wukong & VQDM & BigGAN & \textit{\textbf{Avg.}} \\
\toprule
Joint & 88.4 & 99.7 & 99.6 & 72.5 & 94.4 & 98.6 & 79.4 & 88.9 & 90.2 \\
\rowcolor{yellow!15}
\textbf{Separate} & \textbf{88.8} & \textbf{100} & \textbf{99.9} & \textbf{74.0} & \textbf{95.8} & \textbf{98.9} & \textbf{80.8} & \textbf{89.8} & \textbf{91.0} \\
\bottomrule
\end{tabular}
}
\end{table}

\subsection{Effect of positional encodings in CLS-pooling}
We conduct an ablation to isolate the effect of the positional embeddings when using CLS-pooling. Specifically, we compare the full model (“CLS-pooling w/ pos. enc.”) to a variant that uses the same CLS-based self-attention but omits positional embeddings (“CLS-pooling w/o pos. enc.”), removing any explicit indication of timestep order. As shown in Table~\ref{tab:sequence_order}, providing sequence order information results in a significant improvement of 7.2\% in average accuracy, confirming that timestep position is an important signal when aggregating latents jointly. Despite this gain, the CLS-based variant remains less effective than the default LATTE architecture, which aggregates the outputs via average pooling. Interestingly, the "CLS-pooling w/ pos. enc." variant demonstrates better performance on certain subsets - 9.2\% increase on ADM and 7.9\% on VQDM - suggesting that this CLS-based design, paired with sequence order cues, can be beneficial in specific contexts. 

\vspace{-0.1cm}
\begin{table}[h]
\centering
\caption{\textbf{Accuracy(\%) comparison for CLS-pooling with and without explicit sequence order.} Explicit positional embeddings improve accuracy by 7.2\% over the implicit variant, but fall slightly short of the average pooling.}
\label{tab:sequence_order}
\resizebox{\textwidth}{!}{
\begin{tabular}{ll|ccccccccc}
\toprule
\textbf{Sequence order} & \textbf{Pos. enc.} & Midjourney & SDv1.4 & SDv1.5 & ADM & GLIDE & Wukong & VQDM & BigGAN & \textit{\textbf{Avg.}} \\
\toprule
CLS-pooling & no & 75.5 & 99.7 & 99.6 & 72.2 & 78.9 & 98.3 & 75.6 & 59.3 & 82.4 \\
CLS-pooling & yes & 85.8 & 99.0 & 99.0 & \textbf{83.2} & 86.7 & 96.8 & \textbf{88.7} & 77.8 & 89.6 \\
\midrule
\rowcolor{yellow!15}
\textbf{Avg. pooling} & N/A & \textbf{88.8} & \textbf{100} & \textbf{99.9} & 74.0 & \textbf{95.8} & \textbf{98.9} & 80.8 & \textbf{89.8} & \textbf{91.0} \\
\bottomrule
\end{tabular}
}
\end{table}

\section{Latent Trajectory Spatial Analysis}
\label{appendix:traj_analysis}

To motivate the modeling of the latent trajectory and to distinguish how diffusion-based reconstructions differ between real and generated images, we analyze the spatial distribution of latent denoising corrections across timesteps.

Specifically, we compute the average per-pixel norm of latent differences between consecutive denoising steps - denoted as $\Delta z_t = |z_t - z_{t-1}|_2$ - for a sequence of tracked timesteps ${t_1, t_2, \ldots, t_K}$. For each timestep interval $t_{k-1} \rightarrow t_k$, we aggregate $\Delta z_t$ across all spatial positions and across a batch of samples to obtain a mean spatial correction heatmap:
\begin{equation*}
    H_{t_k}(x, y) = \mathbb{E}_{n}\left[ \left| z_{t_k}^{(n)}(x, y) - z_{t_{k-1}}^{(n)}(x, y) \right|_2 \right],
\end{equation*}

where $(x, y)$ indexes spatial coordinates and $n$ indexes the samples. 
The resulting heatmaps visualize how the latent representation evolves across timesteps by capturing the spatial magnitude of change between consecutive steps. They serve as a proxy for identifying where and how strongly the model updates its latent estimated at each stage of the denoising process. This spatial perspective complements our temporal trajectory modeling and helps reveal structural patterns that distinguish real and generated images. 

\begin{figure}[htbp]
    \centering
    \subfloat[Glide\label{fig:glide}]{
        \includegraphics[width=0.48\textwidth]{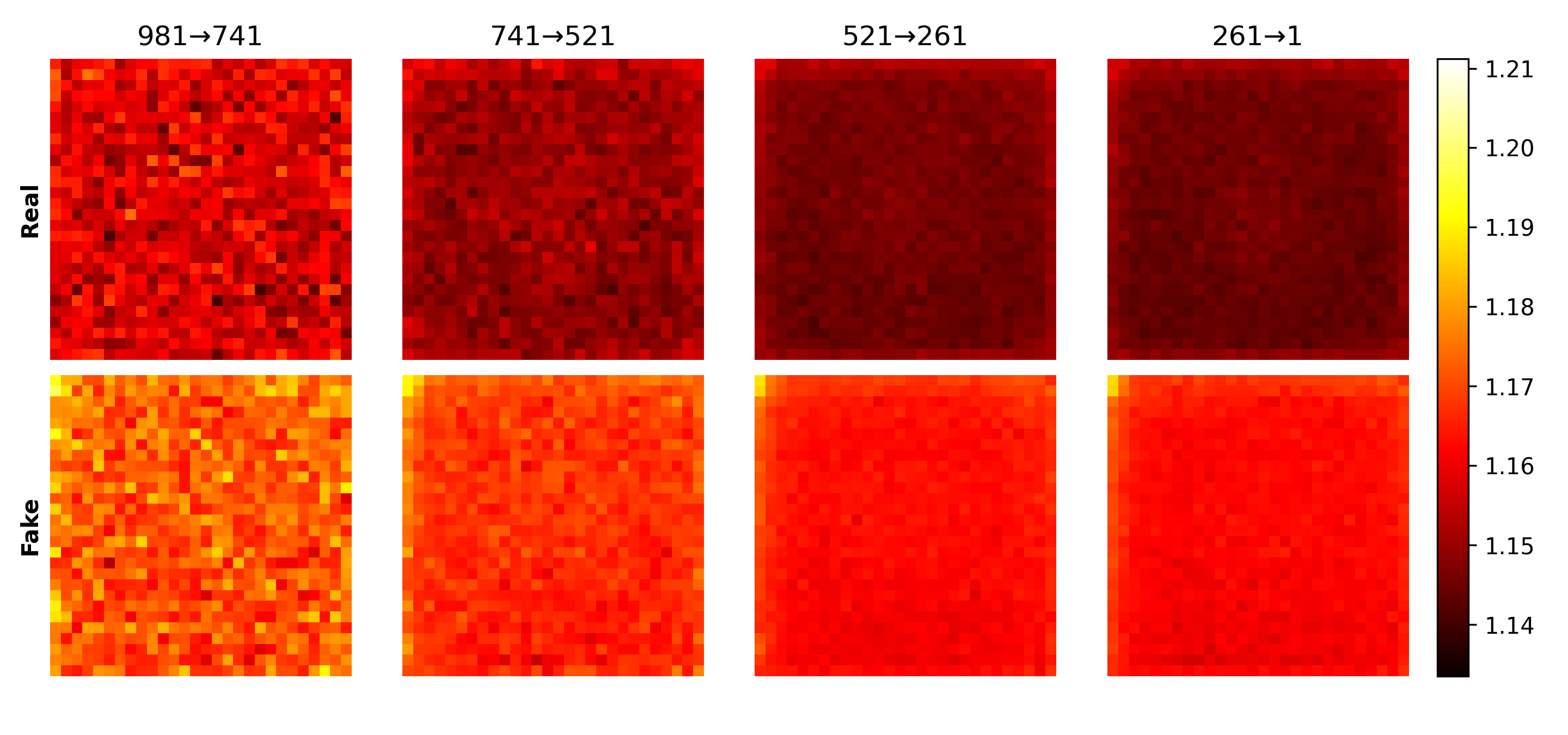}
    }
    \subfloat[ADM\label{fig:adm}]{
        \includegraphics[width=0.48\textwidth]{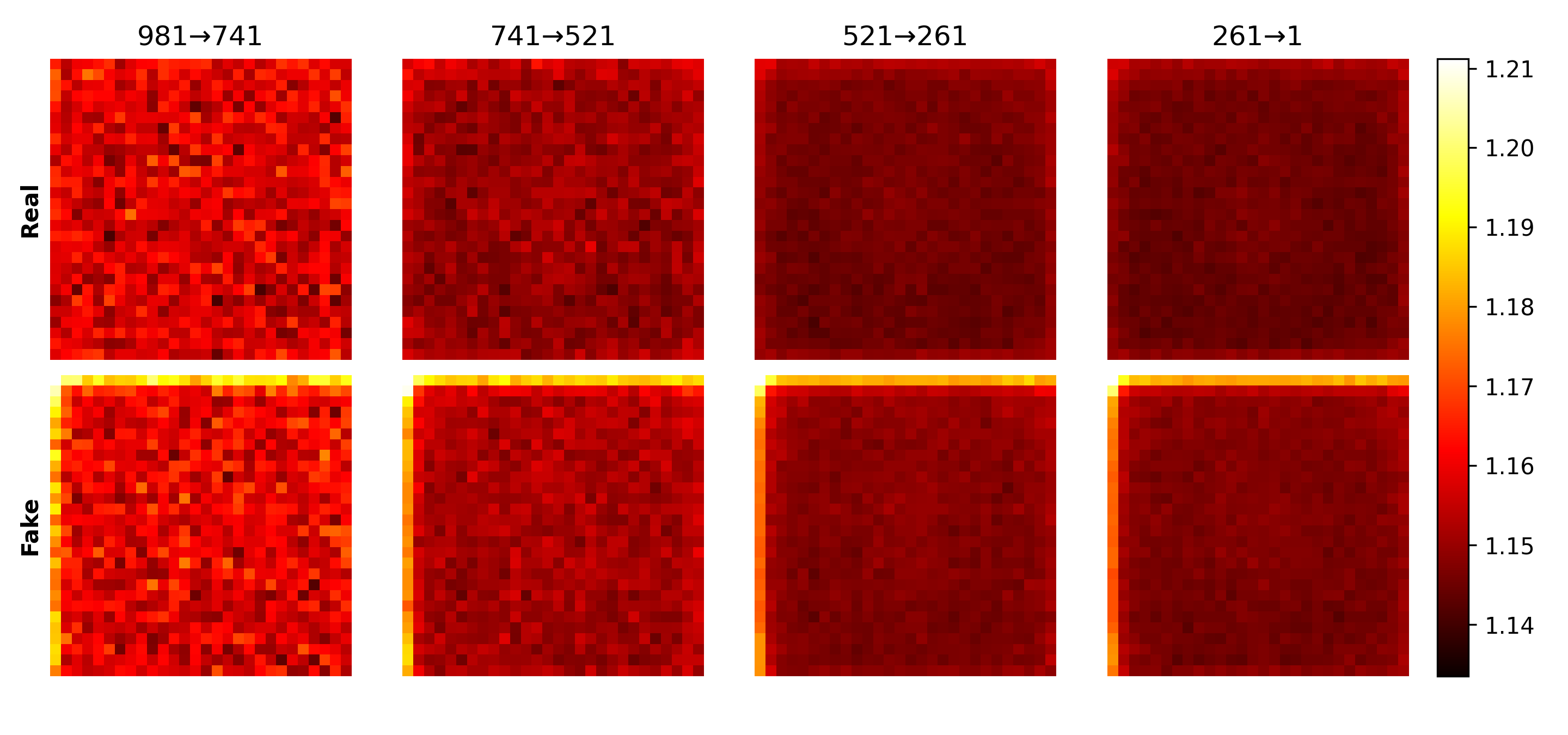}
    }
    \vspace{-1em}
    \subfloat[SDv1.4\label{fig:sdv1.4}]{
        \includegraphics[width=0.48\textwidth]{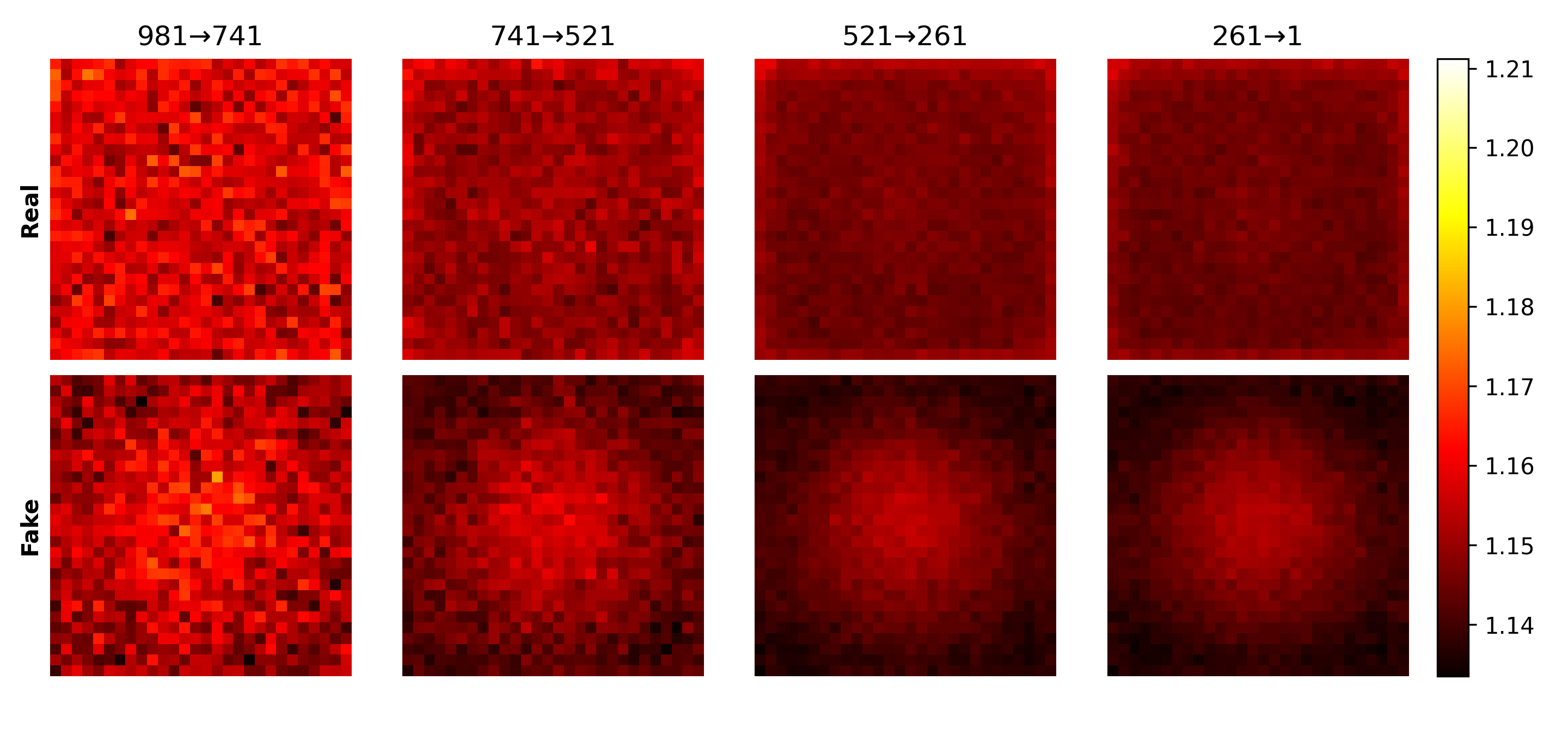}
    }
    \vspace{-1em}
    \hfill
    \subfloat[Midjourney\label{fig:midjourney}]{
        \includegraphics[width=0.48\textwidth]{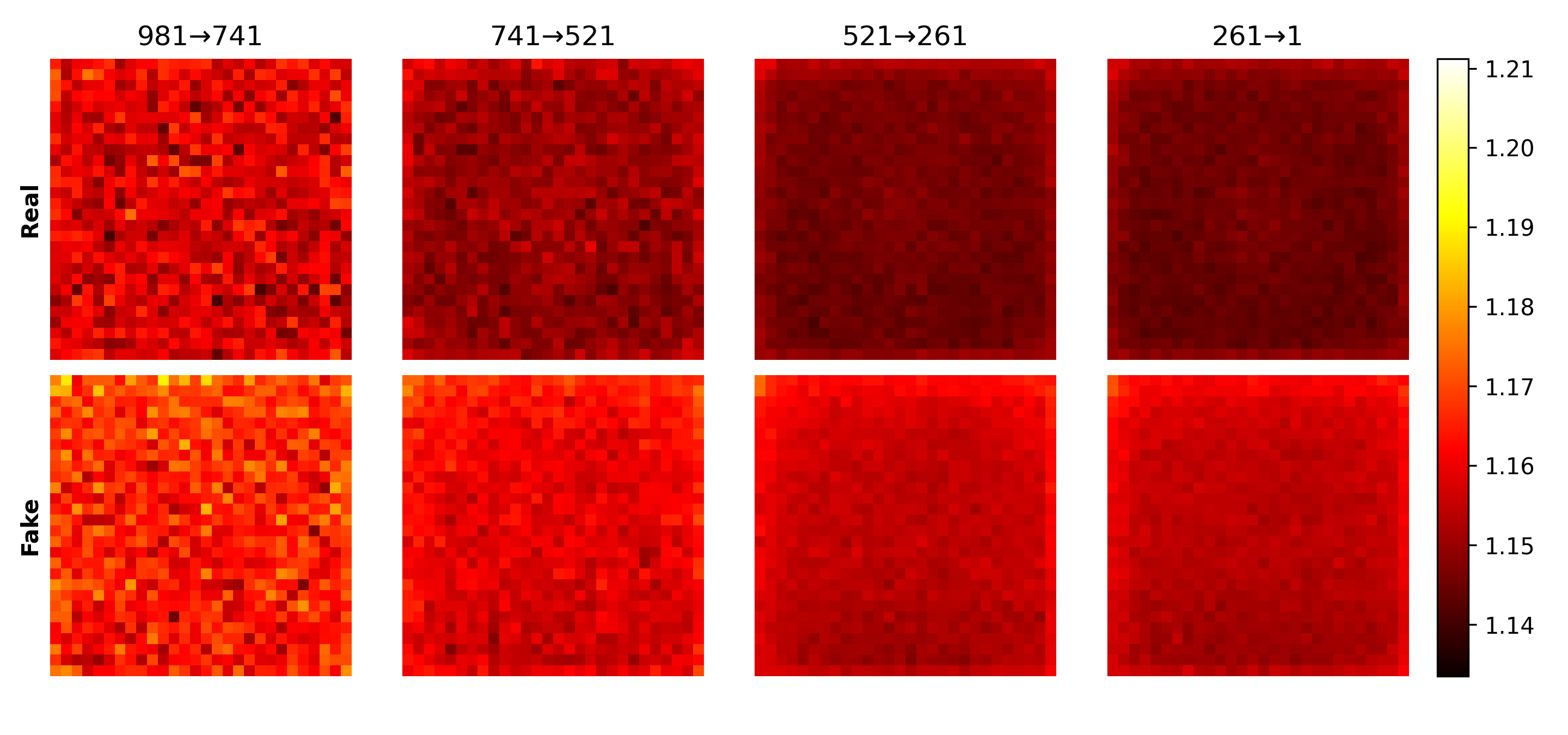}
    }
    \subfloat[BigGAN\label{fig:biggan}]{
        \includegraphics[width=0.48\textwidth]{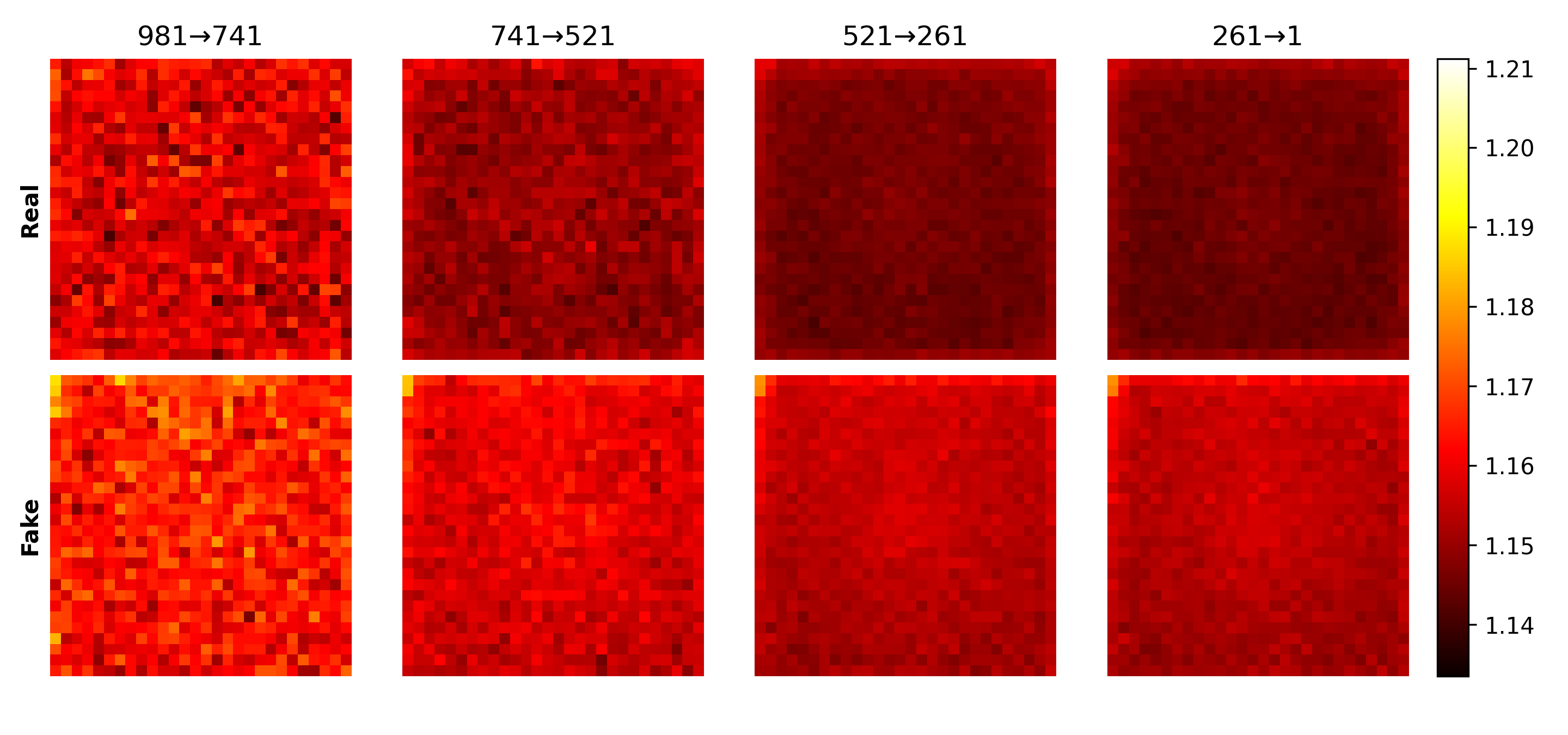}
    }
    \hfill
    \caption{\textbf{Latent trajectory spatial analysis using images from the GenImage dataset.} The real images plots represent averages over all real images in the dataset, while the fakes are plotted separately based on the generators used to produce them.}
    \label{fig:spatial_analysis}
\end{figure}

Based on Figure \ref{fig:spatial_analysis}, we observe a clear dichotomy between real and fake images across most generators. The real images follow a smooth, uniformly paced denoising trajectory, indicating that each denoising correction is modest in magnitude and spatially consistent. 

Fake images, in contrast, break this steady pattern in different ways. Images generated by GLIDE (\ref{fig:glide}) require substantially larger corrections overall. The early steps are especially bright - indicating heavier refinement in the beginning of the denoising process - before tapering off into smaller updates. Midjourney (\ref{fig:midjourney}) and BigGAN (\ref{fig:biggan}) behave almost identically, with lower differences between real and fake heatmaps than Glide, but still pronounced at every step. Unlike the real's constant gradual decline, their fake trajectories show a striking front-loaded burst: the jump in $\Delta z$ between the first two steps is far greater than any subsequent change. This pattern reveals that, for these generators, most of the refinement occurs in the first half of the trajectory, with little correction applied later.

By contrast, the ADM subset (\ref{fig:adm}) shows a markedly different trend. Here, the real vs. fake differences across all steps are considerably more subtle, and the resulting $\Delta z$ heatmaps for both classes appear visually similar in both magnitude and spatial pattern, with the exception of small brighter left and top margins. This behavior is consistent with our model's relatively poor performance on ADM (74\% compared to the 91\% average) and suggests that the images in this subset are particularly difficult to distinguish - even in trajectory space.

Finally, SDv1.4 (\ref{fig:sdv1.4}) presents the most distinctive behavior. Unlike previous generators, the fake heatmaps exhibit a center-focused $\Delta z$ signature. This effect likely arises because we use Stable Diffusion \citep{sdv} for both generating and reconstructing the images. The denoiser has learned to prioritize central content - where objects are typically located during prompt-guided generation - and thus applies larger, spatially focused corrections in the center of the image. Real images, by contrast, lack this learned structure and receive relatively uniform and lower-magnitude corrections across space.

\section{Complete accuracy \& average precision on GenImage}
\label{appendix:complete_acc}

\begin{figure}[h]
  \centering
  \begin{minipage}{0.49\textwidth}
    \centering
    \includegraphics[width=\linewidth]{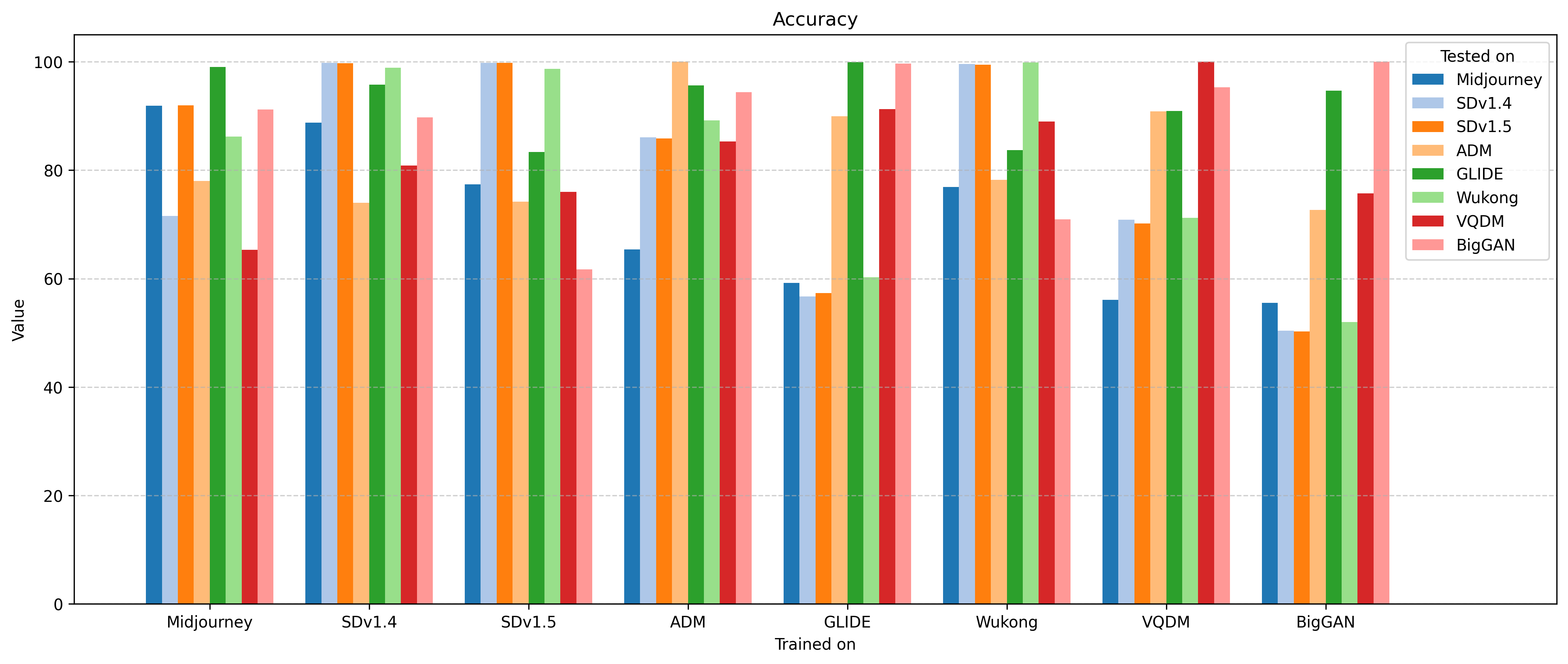}
    \label{fig:image1}
  \end{minipage}\hfill
  \begin{minipage}{0.49\textwidth}
    \centering
    \includegraphics[width=\linewidth]{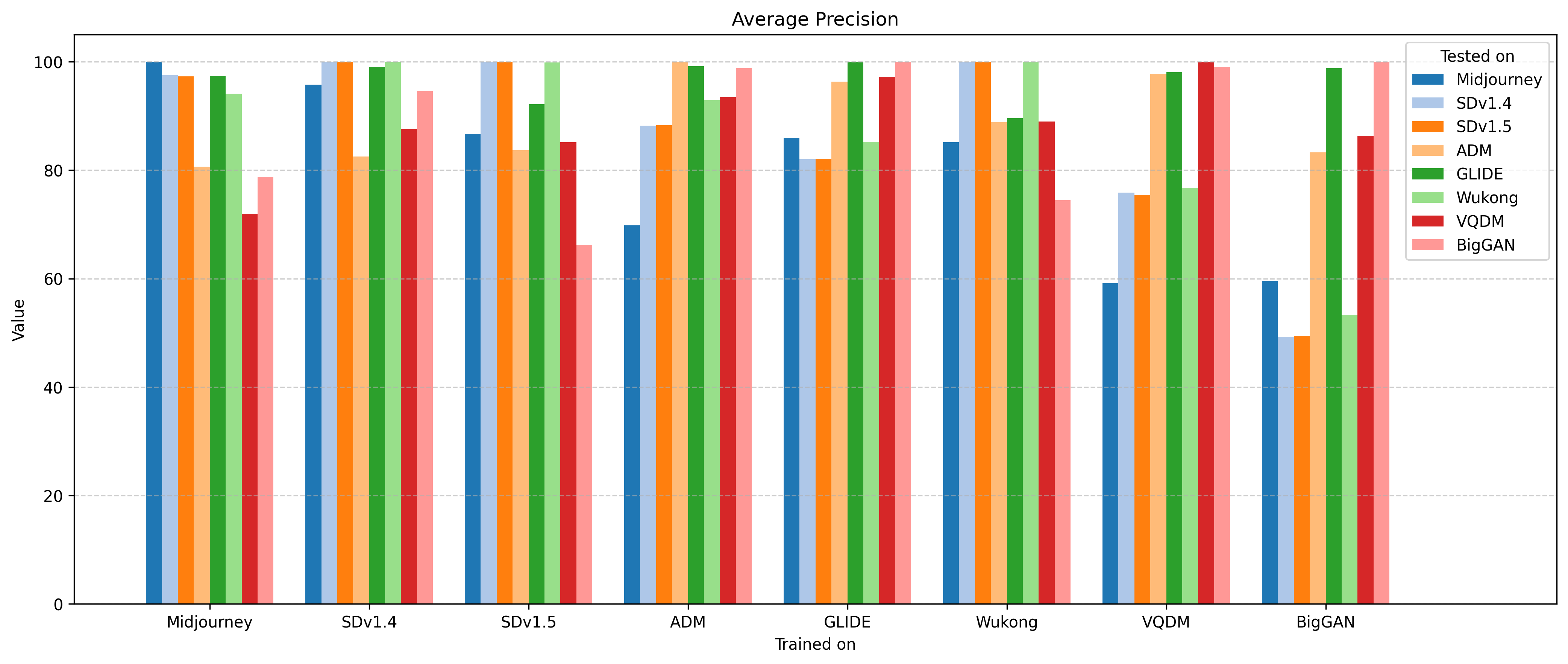}
    \label{fig:image2}
  \end{minipage}
  \vspace{-0.5cm}
  \caption{\textbf{Accuracy(\%) (left) and average precision(\%) (right) of LATTE across the GenImage dataset.} The x-axis indicates the generator used to produce the training data, while each bar represents the model's performance when tested on data from the different generators. }
  \label{fig:genimage_complete}
\end{figure}

Figure \ref{fig:genimage_complete} presents LATTE’s performance on the GenImage dataset, reporting both accuracy and average precision across different training–testing generator combinations. The results show that LATTE maintains consistently high performance regardless of the generator used for training, highlighting its ability to generalize across diverse generative models. 

\section{Architectural Details of the CLS-pooling}
\label{appendix:cls_pooling}

\begin{figure}[h]
    \centering
    \includegraphics[width=0.95\textwidth]{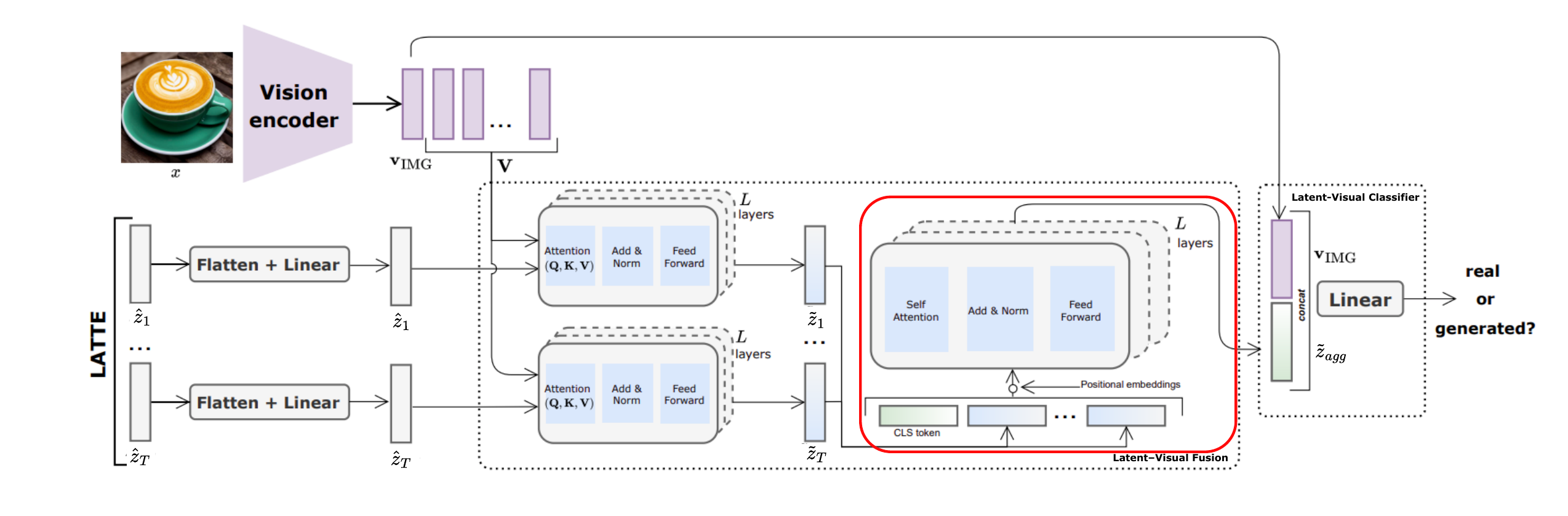}
     \vspace{-0.25cm}
    \caption{\textbf{Overview of our proposed LATTE architecture with CLS pooling as an aggregation strategy (denoted in red).} A learnable CLS token is prepended to the fused sequence and processed via a self-attention stack.}
    \label{fig:latte_cls}
\end{figure}

We consider CLS-pooling as an alternative aggregation strategy (instead of average pooling), illustrated in Figure \ref{fig:latte_cls}. After independently fusing each projected latent $\tilde z_{t}$ with visual features, through a stack of transformer decoder layers, a learnable token $z_{\text{CLS}} \in \mathbb{R}^d$ is prepended to the sequence of refined latent embeddings $\widetilde{\mathcal{T}}(x) = \{\tilde z_{t_1}, \tilde z_{2}, \dots, \tilde z_{t_K}\}$. Learnable positional embeddings are added to this sequence to inform the model of the order of timesteps. The sequence is then passed through a shared self-attention stack of transformer layers, allowing the CLS token $z_{\text{CLS}}$ to interact with the full latent trajectory and aggregate information across timesteps. The final CLS token output serves as the aggregated trajectory representation $\tilde z_{\text{agg}}$. The rest of the architecture remains the same as in Figure \ref{fig:latte_architecture}.

\section{Embedding Space Analysis}
\label{appendix:embddings}

\begin{figure}[h]
    \centering
    \includegraphics[width=0.7\textwidth]{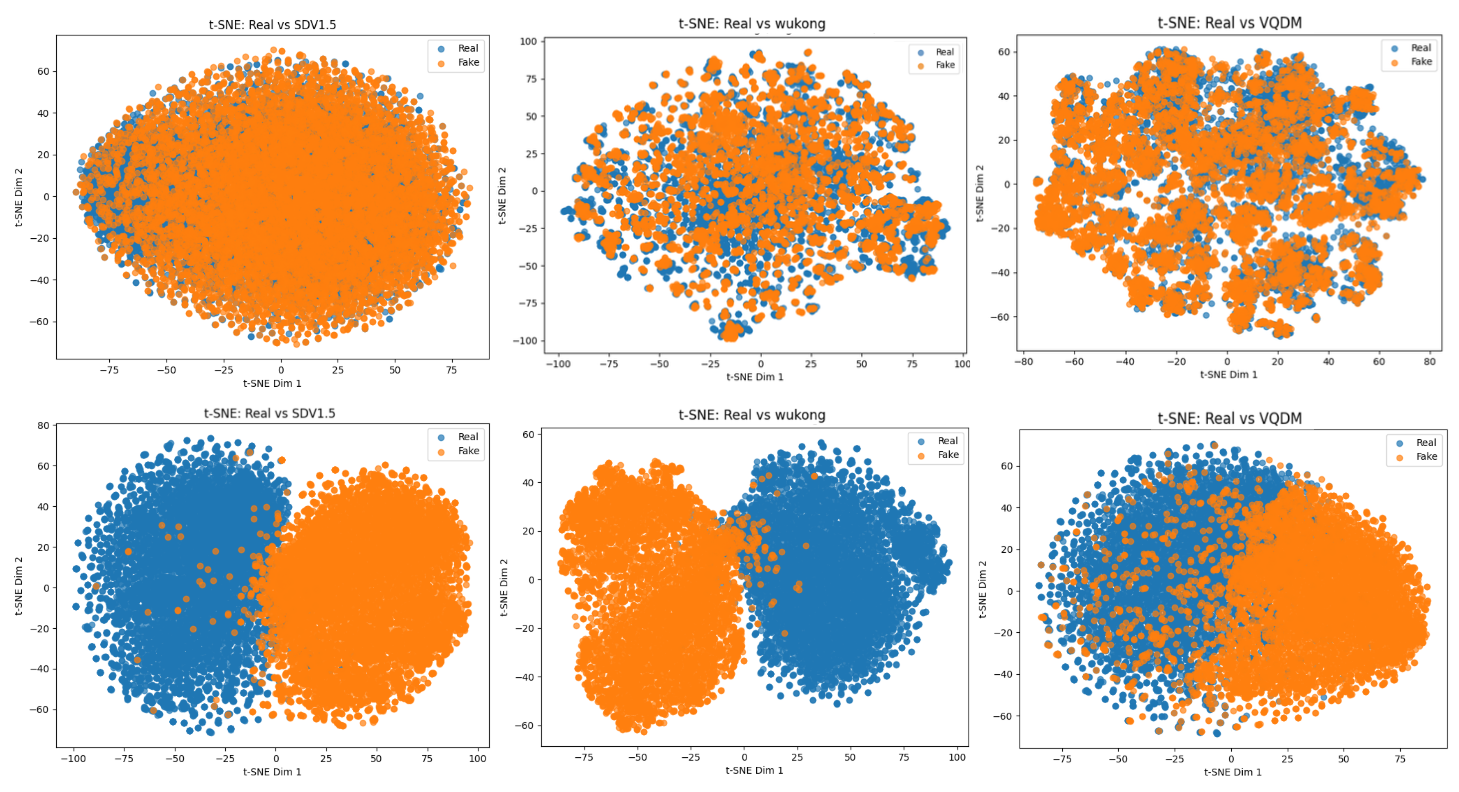}
     \vspace{-0.25cm}
    \caption{\textbf{Visualizations of t-SNE embeddings for real and fake images across the remaining three generators from GenImage.} The first row presents embeddings before using LATTE (extracted using the original ConvNeXt), while the second row shows embeddings derived from LATTE. }
    \label{fig:embeddings_remaining}
\end{figure}

To complete our embedding space analysis from Section 4.4, Figure \ref{fig:embeddings_remaining} presents t-SNE plots for the three remaining subsets in the GenImage dataset, namely the SDv1.5 \citep{sdv}, Wukong \citep{wukong}, and VQDM \citep{vqdm} generators. As in Figure \ref{fig:embeddings}, the top row shows embeddings extracted with the frozen ConvNeXt backbone \citep{convnext} (pre-LATTE) and the bottom row shows embeddings after LATTE fine-tuning. The much clearer separation in the second row illustrates LATTE's discriminative power.

\end{document}